\documentclass{tlp}

\submitted{23 May 2010}
\revised{~} 
\accepted{18 August 2010}

\usepackage{amsmath}

\def\ar{\leftarrow}
\def\beq{\begin{equation}}
\def\eeq#1{\label{#1}\end{equation}}
\def\ba{\begin{array}}
\def\ea{\end{array}}

\def\no{\it not\/}
\def\ar{\leftarrow}
\def\rar{\rightarrow}

\def\proof{\noindent{\bf Proof}.\hspace{3mm}}
\def\qed{\quad \vrule height7.5pt width4.17pt depth0pt \medskip}

\newcounter{mythmc}

\newtheorem{prop}{Proposition}
\newtheorem{thm}{Theorem}
\newtheorem{mythm}[mythmc]{Theorem}
\newtheorem{mythmd}[mythmc]{Theorem$^{\rm d}$}
\newtheorem{mythme}[mythmc]{Theorem$^{\rm hef}$}
\newtheorem{mypropd}[mythmc]{Proposition$^{\rm d}$}
\newtheorem{mycold}[mythmc]{Corollary$^{\rm d}$}
\newtheorem{cor}{Corollary}

\newtheorem{definition}{Definition}

\def\Pish{\Pi_\mathit{sh}}

\begin{document}

\title[On Elementary Loops of Logic Programs]{\bf On Elementary Loops
  of Logic Programs}
\author[Martin~Gebser \and Joohyung Lee \and Yuliya Lierler]{
Martin Gebser \\
Institut f\"ur Informatik \\
Universit\"at Potsdam, Germany \\
\email{gebser@cs.uni-potsdam.de}
\and 
Joohyung Lee \\ 
School of Computing, Informatics and Decision Systems Engineering \\
Arizona State University, USA \\
\email{joolee@asu.edu}
\and
Yuliya Lierler \\
Department of Computer Science \\
University of Kentucky, USA \\
\email{yuliya@cs.uky.edu}
}


\maketitle

\noindent
{\bf Note:} To appear in {\sl Theory and Practice of Logic Programming (TPLP)}
\bigskip


\begin{abstract}
Using the notion of an elementary loop, Gebser and Schaub refined the
theorem on loop formulas due to Lin and Zhao by considering loop
formulas of elementary loops only. In this article, we reformulate
their definition of an elementary loop, extend it to disjunctive
programs, and study several properties of elementary loops, including
how maximal elementary loops are related to minimal unfounded
sets. The results provide useful insights into the stable model
semantics in terms of elementary loops. For a nondisjunctive program,
using a graph-theoretic characterization of an elementary loop, we
show that the problem of recognizing an elementary loop is
tractable. On the other hand, we show that the corresponding problem
is {\sf coNP}-complete for a disjunctive program. Based on the notion
of an elementary loop, we present the class of
Head-Elementary-loop-Free (HEF) programs, which strictly generalizes
the class of Head-Cycle-Free (HCF) programs due to Ben-Eliyahu and
Dechter. Like an HCF program, an HEF program can be turned into
an equivalent nondisjunctive program in polynomial time by shifting
head atoms into the body.
\end{abstract}

\section{Introduction} \label{sec:intro}

The theorem on loop formulas due to Lin and Zhao \citeyear{lin04} has
contributed to understanding the relationship between the stable model
semantics and classical logic. Unlike other translations that modify the
vocabulary of a logic program \cite{ben94,lin03,janhunen06a}, the
original theorem on loop formulas characterizes the stable models of a
nondisjunctive program in terms of the models of its completion that
satisfy the loop formulas of all loops of the program. This allows us
to compute stable models using SAT solvers, which led to the
design of answer set solvers
{\sc assat}\footnote{{\tt http://assat.cs.ust.hk/}} \cite{lin04}
and 
{\sc cmodels}\footnote{{\tt http://www.cs.utexas.edu/users/tag/cmodels/}}
\cite{giu04a}.
Due to its importance in semantic understanding as well as in stable
model computation, the theorem on loop formulas has been extended to
more general classes of logic programs, such as disjunctive programs
\cite{lee03a}, infinite programs and programs containing
classical negation \cite{lee05}, and programs containing
aggregates \cite{liul06,youj08,leej09a}. Moreover, it has been applied
to other nonmonotonic logics, such as circumscription \cite{lee06} and
nonmonotonic causal logic \cite{lee03b}.
The stable model semantics for first-order formulas given in
\cite{fer07a,ferr09} is also closely related to the idea of loop
formulas, as described in \cite{lee08a}. 

By slightly modifying the definition of a loop, Lee~\citeyear{lee05}
showed that loop formulas can be viewed as a generalization of
completion~\cite{cla78}. The model-theoretic account of loop formulas
give in~\cite{lee05} also tells us that the idea of loop formulas is
closely related to assumption sets~\cite{sac90} or unfounded
sets~\cite{leo97}. In a sense, the theorem by Lin and Zhao is an
enhancement of the unfounded set based characterization of stable
models given in~\cite{sac90,leo97}. The unfounded set based
characterization takes into account the loop formulas of {\sl all sets
  of atoms}, while the theorem by Lin and Zhao considers the loop
formulas of {\sl loops} only. 
Gebser and Schaub \citeyear{gebsch05a} improved this enhancement even
further. They defined the notion of an elementary loop of a
nondisjunctive program and showed that the theorem by Lin and Zhao
remains correct even if we consider loop formulas of {\sl elementary
  loops} only. 

In this article, we reformulate the definition of an elementary loop
of a nondisjunctive program by Gebser and Schaub, extend it to
disjunctive programs, and study several properties of elementary
loops, including how maximal elementary loops are related to minimal
unfounded sets. 
Based on the notion of an elementary loop, we present the class of 
Head-Elementary-loop-Free (HEF) program, which strictly generalizes 
the class of Head-Cycle-Free (HCF) programs due to Ben-Eliyahu and
Dechter \citeyear{ben94}. Like an HCF program, an HEF program can be
turned into an equivalent nondisjunctive program in polynomial time by
shifting head atoms into the body---a simple transformation defined in
\cite{gel91a}. This tells us that an HEF program is an ``easy''
disjunctive program, which is merely a syntactic variant of a
nondisjunctive program. 
We also observe that several other properties of nondisjunctive
and HCF programs can be generalized to HEF programs.
The main results from~\cite{lin03} and \cite{you03}, characterizing
stable models in terms of {\em inherent tightness} and {\em weak
  tightness}, respectively, can be extended to HEF programs, and
likewise the operational characterization of stable models of HCF
programs due to Leone {\sl et al.} \citeyear{leo97} can be extended to
HEF programs. 
The properties of elementary loops and HEF programs studied here may
be useful in improving the computation of disjunctive answer set
solvers, such as 
{\sc claspd}\footnote{{\tt http://potassco.sourceforge.net/}}
\cite{drgegrkakoossc08a},
{\sc cmodels}
\cite{lie05},
{\sc dlv}\footnote{{\tt http://www.dbai.tuwien.ac.at/proj/dlv/}}
\cite{leo06a}, and 
{\sc gnt}\footnote{{\tt http://www.tcs.hut.fi/Software/gnt/}}
\cite{jan06}.

The outline of this paper is as follows. In
Section~\ref{sec:elm-nondis}, we present our reformulated definition
of an elementary loop of a nondisjunctive program and provide a
corresponding refinement of the theorem on loop formulas, 
as well as some properties of elementary loops. These results are
extended to disjunctive programs in Section~\ref{sec:elm-dis}. 
In Section~\ref{sec:hef}, we introduce the class of HEF programs
and show that their shifted variants preserve stable models. In
Section~\ref{sec:tight}, we generalize the notion of inherent
tightness to HEF programs.
An operational characterization of stable models of HEF programs
is presented in Section~\ref{sec:msv}.
Finally, Section~\ref{sec:conclusion} concludes the paper.

This paper is an extended version of the conference papers
\cite{gll06} and \cite{gll07}.\footnote{%
In \cite{gll06,gll07}, the term  ``elementary set'' was used in place
of ``elementary loop.''}

\section{Nondisjunctive Programs}\label{sec:elm-nondis}

After providing the relevant background on nondisjunctive programs,
this section introduces elementary loops of nondisjunctive programs.
We further refine elementary loops to elementarily unfounded
sets, yielding a syntactic characterization of minimal unfounded sets.
Moreover, we show that elementary loops of nondisjunctive programs
can be recognized in polynomial time. (The statements of the theorems
and the propositions in this section which apply to nondisjunctive
programs will be generalized to disjunctive programs or HEF programs
in later sections and the proofs will be given there.)
Finally, we compare our reformulation of elementary
loops with the definition by Gebser and Schaub \citeyear{gebsch05a}.

\subsection{Background} 

A {\sl nondisjunctive rule} is an expression of the form
\beq
  a_1\, \ar\, a_2,\dots,a_m,\no\ a_{m+1},\dots,\no\ a_n
\eeq{nd}
where $n\ge m\ge 1$ and $a_1,\dots,a_n$ are propositional atoms. A
{\sl nondisjunctive program} is a finite set of nondisjunctive rules.  

We will identify a nondisjunctive rule~(\ref{nd}) with the
propositional formula
\beq 
  a_2\land\cdots\land a_m\land\neg a_{m+1}\land\cdots\land\neg a_n
  \, \rar\, a_1\ , 
\eeq{nd-form} 
and will often write~(\ref{nd}) as
\beq
  a_1\, \ar\, B, F
\eeq{abf-nd}
where $B$ is $a_2,\dots,a_m$ and $F$ is $\no\ a_{m+1},\dots,\no\ a_n$. 
We will sometimes identify~$B$ with its corresponding set of atoms.

We will identify an interpretation with the set of atoms that are true
in it. We say that a set $X$ of atoms satisfies a rule~(\ref{nd}) if
$X$ satisfies (\ref{nd-form}).
Moreover, $X$ satisfies a nondisjunctive program $\Pi$ (symbolically,
$X\models\Pi$) if $X$ satisfies every rule~(\ref{nd}) of~$\Pi$. 
If $X$ satisfies~$\Pi$, we also call $X$ a {\sl model} of~$\Pi$.

The reduct $\Pi^X$ of a nondisjunctive program $\Pi$ w.r.t. a set $X$
of atoms is obtained from $\Pi$ by deleting each rule~(\ref{abf-nd}) 
such that $X\not\models F$, and replacing each remaining
rule~(\ref{abf-nd}) with $a_1\ar B$. A set $X$ of atoms is a {\sl
stable model}, also called an {\sl answer set}, of $\Pi$ if $X$ is
minimal among the sets of atoms that satisfy $\Pi^X$.

The {\sl (positive) dependency graph} of a nondisjunctive
program~$\Pi$ is the directed graph such that its vertices are the
atoms occurring in~$\Pi$, and its edges go from~$a_1$ to~$a_2,\dots,
a_m$ for all rules~(\ref{nd}) of~$\Pi$.
A nonempty set~$Y$ of atoms is called a {\sl loop} of~$\Pi$ if, for every 
pair~$a$, $b$ of atoms in~$Y$, there is a path (possibly of length~$0$)
from~$a$ to $b$ in the dependency graph of~$\Pi$ such that all vertices 
in the path belong to~$Y$.  
In other words, a nonempty set~$Y$ of atoms that occur in~$\Pi$ is a
loop of~$\Pi$ if the subgraph of the dependency graph of~$\Pi$ induced
by~$Y$ is strongly connected.  
It is clear that every singleton whose atom occurs in~$\Pi$ is a loop
of~$\Pi$.

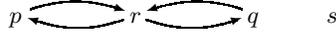
\begin{figure}
\begin{center}
 \begin{picture}(120,10)(0,3)
 \put(0,0){$p$}
 \put(90,0){$q$}
 \put(45.5,0){$r$}
 \put(120,0){$s$}

 \qbezier(8,5)(25,11)(42,5) \put(43,4.6){\vector(3,-1){1}}
 \qbezier(43,1)(26,-5)(9,1) \put(8,1.3){\vector(-3,1){1}}

 \qbezier(88,5)(71,11)(54,5) \put(53,4.6){\vector(-3,-1){1}}
 \qbezier(52,1)(69,-5)(86,1) \put(87,1.3){\vector(3,1){1}}
 \end{picture}
\caption{The dependency graph of Program~$\Pi_1$}
\label{fig:dep1}
\end{center}
\end{figure}

For illustration, consider the following program $\Pi_1$:
\[ 
\ba {l}
  p \ar \no\ s \\
  p \ar r \\
  q \ar r \\
  r \ar p,q  \ .  
\ea
\] 
Figure~\ref{fig:dep1} shows the dependency graph of $\Pi_1$.
Program~$\Pi_1$ has seven loops: $\{p\}$, $\{q\}$, $\{r\}$, $\{s\}$,
$\{p,r\}$, $\{q,r\}$, and $\{p,q,r\}$.

For any set $Y$ of atoms, the {\sl external support formula} of~$Y$
for a nondisjunctive program~$\Pi$, denoted by ${\it ES}_\Pi(Y)$, is the
disjunction of conjunctions $B\land F$ for all rules~(\ref{abf-nd})
of~$\Pi$ such that $a_1\in Y$ and  $B\cap Y = \emptyset$.
The first condition expresses that the atom ``supported''
by~(\ref{abf-nd}) is an element of~$Y$.  The second condition ensures
that this support is ``external'': the atoms in~$B$ that it relies on
do not belong to~$Y$. Thus, $Y$ is called {\sl externally supported}
by $\Pi$ w.r.t. a set~$X$ of atoms if $X\models{\it ES}_\Pi(Y)$.

For any set $Y$ of atoms, by ${\it LF}_\Pi(Y)$, we denote the following 
formula:
\beq
   \bigwedge_{a\in Y} a\ \rar\ {\it ES}_\Pi(Y) \ .
\eeq{lf}
(The expression in the antecedent stands for the conjunction of all
elements in $Y$.)
Formula~(\ref{lf}) is called the {\sl (conjunctive) loop formula} of~$Y$
for~$\Pi$. Note that we still call (\ref{lf}) a loop formula even when
$Y$ is not a loop of~$\Pi$.

The following reformulation of the Lin-Zhao theorem, which
characterizes the stability of a model by loop formulas, is a part of
the main theorem from \cite{lee05} for the nondisjunctive case.

\begin{thm}[\cite{lee05}] \label{thm:lf}
For any nondisjunctive program~$\Pi$ and any set~$X$ of atoms that
occur in~$\Pi$, if $X$ is a model of~$\Pi$, then the following
conditions are equivalent: 
\begin{itemize}
\item[(a)]  $X$ is a stable model of~$\Pi$;
\item[(b)]  $X$ satisfies~${\it LF}_\Pi(Y)$ for all nonempty sets $Y$ 
            of atoms that occur in~$\Pi$;            
\item[(c)]  $X$ satisfies~${\it LF}_\Pi(Y)$ for all loops $Y$ of~$\Pi$.
\end{itemize}
\end{thm}

According to the equivalence between conditions~(a) and~(b) in
Theorem~\ref{thm:lf}, a model of~$\Pi_1$ is stable iff it satisfies 
the loop formulas of all fifteen nonempty sets of atoms formed from
the atoms occurring in~$\Pi_1$. 
On the other hand, condition~(c) tells us that it is sufficient to
restrict attention to the following loop formulas of the seven loops
of~$\Pi_1$:
\beq
\ba l 
   p \rar \neg s \lor r \\
   q \rar  r \\
   r \rar  p \land q \\
   s \rar \bot \\
   p \land r \rar \neg s  \\
   q \land r \rar \bot \\
   p \land q \land r \rar \neg s \ . 
\ea
\eeq{ex-lf} 
Program~$\Pi_1$ has six models: $\{p\}$, $\{s\}$, $\{p,s\}$,
$\{q,s\}$, $\{p,q,r\}$, and $\{p,q,r,s\}$. Among them, $\{p\}$ is the
only stable model of~$\Pi_1$, which is also the only model of~$\Pi_1$
that satisfies all loop formulas in~(\ref{ex-lf}). 

As noted in \cite{lee05}, the equivalence between conditions~(a) and
(c) is a reformulation of the theorem by Lin and Zhao;  the equivalence
between conditions~(a) and (b) is a reformulation of Corollary~2 from
\cite{sac90} and Theorem~4.6 from \cite{leo97} (in the nondisjunctive
case), which characterizes the stability of a model in terms of
unfounded sets. 
For any sets $X$, $Y$ of atoms, we say that $Y$ is {\sl unfounded}
by~$\Pi$ w.r.t.~$X$ if $Y$ is not externally supported by~$\Pi$
w.r.t.~$X$. Condition~(b) can be stated in terms of unfounded sets as
follows:
\begin{itemize}
\item[(b$'$)] $X$ contains no nonempty unfounded sets 
              for~$\Pi$ w.r.t.~$X$.
\end{itemize}


\subsection{Elementary Loops of Nondisjunctive Programs}\label{ssec:elm}

Gebser and Schaub~\citeyear{gebsch05a} showed that $Y$ in
${\it LF}_\Pi(Y)$ in Theorem~\ref{thm:lf} can be restricted to
``elementary'' loops only. In this section, we present a reformulation
of their definition of an elementary loop and investigate its
properties. We compare our reformulation with the original definition
by Gebser and Schaub in Section~\ref{sec:comparison}.

To begin with, the following proposition tells us that a loop can be
defined without mentioning a dependency graph.

\begin{prop}\label{prop:loop-alternative}
For any nondisjunctive program~$\Pi$ and any nonempty set~$X$ of atoms
that occur in~$\Pi$, $X$ is a loop of~$\Pi$ iff, for every nonempty
proper subset~$Y$ of $X$, there is a rule~(\ref{abf-nd}) in~$\Pi$ such
that $a_1\in Y$ and $B\cap (X\setminus Y)\ne\emptyset$.
\end{prop}

For any set $X$ of atoms and any subset~$Y$ of~$X$, we say that 
$Y$ is {\sl outbound} in~$X$ for a nondisjunctive program~$\Pi$ 
if there is a rule~(\ref{abf-nd}) in~$\Pi$ such that 
\begin{itemize}
\item $a_1\in Y$, 
\item $B\cap (X\setminus Y)\ne\emptyset$, and
\item $B\cap Y=\emptyset$.
\end{itemize}

The following proposition describes the relationship between the
external support formula of a set~$Y$ of atoms and the external
support formula of a subset~$Z$ of~$Y$ that is not outbound in~$Y$. 

\begin{prop}\label{prop:es}
For any nondisjunctive program $\Pi$ and any sets $X$, $Y$, $Z$ of
atoms such that \hbox{$Z\subseteq Y\subseteq X$}, if~$Z$ is not
outbound in~$Y$ for~$\Pi$ and $X\models{\it ES}_\Pi(Z)$, then
$X\models{\it ES}_\Pi(Y)$.
\end{prop}

Proposition~\ref{prop:es} tells us that, in order to verify that a
set~$Y$ of atoms is externally supported by~$\Pi$ w.r.t. a
superset~$X$ of~$Y$, it is sufficient to identify some externally
supported subset of~$Y$ that is not outbound in~$Y$ for~$\Pi$.
Conversely, if $Y$ is not externally supported by $\Pi$ w.r.t.~$X$,
then every subset of $Y$ that is externally supported by $\Pi$
w.r.t.~$X$ is outbound in $Y$ for $\Pi$.

For any nonempty set $X$ of atoms that occur in~$\Pi$, we say that~$X$
is an {\sl elementary loop} of~$\Pi$ if all nonempty proper subsets
of~$X$ are outbound in~$X$ for~$\Pi$. As with loops, it is clear from
the definition that every singleton whose atom occurs in~$\Pi$ is an
elementary loop of~$\Pi$. It is also clear that every elementary loop
of~$\Pi$ is a loop of $\Pi$: the condition for being an elementary
loop implies the condition for being a loop as stated in
Proposition~\ref{prop:loop-alternative}.
On the other hand, a loop is not necessarily an elementary loop.
For instance, one can check that $\{p,q,r\}$  is not an elementary
loop of~$\Pi_1$ since $\{p,r\}$ (or $\{q,r\}$) is not outbound in
$\{p,q,r\}$ for~$\Pi_1$.  All other loops of~$\Pi_1$ are elementary
loops.
Note that an elementary loop may be a proper subset of another
elementary loop (both $\{p\}$ and $\{p,r\}$ are elementary loops
of~$\Pi_1$). 

The following program replaces the last rule of~$\Pi_1$ with two other
rules:
\beq 
\ba l 
  p \ar \no\ s \\
  p \ar r \\
  q \ar r \\
  r \ar p \\
  r \ar q \ .
\ea
\eeq{ex2}
The program has the same dependency graph as~$\Pi_1$, and hence has
the same loops. However, its elementary loops are different from those
of~$\Pi_1$: all its loops are elementary loops as well, including
$\{p,q,r\}$. 

The definition of an elementary loop~$X$ given above is not affected
if we check the outboundness condition only for all loops or for all
elementary loops that belong to $X$ instead of all nonempty proper
subsets of~$X$. 

\begin{prop}\label{prop:elm-alt}
For any nondisjunctive program~$\Pi$ and any nonempty set $X$ of atoms that
occur in~$\Pi$, $X$ is an elementary loop of~$\Pi$ iff all proper
subsets of $X$ that are elementary loops of~$\Pi$ are outbound in $X$
for $\Pi$.
\end{prop}

The following proposition describes an important relationship between
loop formulas of elementary loops and loop formulas of arbitrary sets
of atoms.

\begin{prop}\label{prop:elm-lf-entail}
For any nondisjunctive program $\Pi$ and any nonempty set $Y$ of atoms
that occur in~$\Pi$, there is an elementary loop $Z$ of $\Pi$ such
that $Z$ is a subset of $Y$ and ${\it LF}_\Pi(Z)$ entails ${\it LF}_\Pi(Y)$.
\end{prop}

Proposition~\ref{prop:elm-lf-entail} allows us to limit attention to
loop formulas of elementary loops only. This yields the following
theorem, which is a reformulation of Theorem~3 from~\cite{gebsch05a}.

\setcounter{mythmc}{1}
\addtocounter{mythmc}{-1}
\begin{mythm}[d]
The following condition is equivalent to each of conditions~(a)--(c) in 
Theorem~\ref{thm:lf}:
\begin{itemize}
\item[(d)]  $X$ satisfies~${\it LF}_\Pi(Y)$ for all elementary 
            loops~$Y$ of~$\Pi$. 
\end{itemize}
\end{mythm}

For instance, according to Theorem~\ref{thm:lf}~(d), a model of~$\Pi_1$
is stable iff it satisfies the first six loop formulas in~(\ref{ex-lf});
the loop formula of the non-elementary loop $\{p,q,r\}$ (the last one
in~(\ref{ex-lf})) can be disregarded.


\subsection{Elementarily Unfounded Sets for Nondisjunctive Programs} 

If we modify condition~(c) in Theorem~\ref{thm:lf} by replacing
``loops'' in its statement with ``maximal loops,'' the condition
becomes weaker, and the modified statement of Theorem~\ref{thm:lf} 
is incorrect. For instance, $\Pi_1$ has only two maximal loops, 
$\{p,q,r\}$ and $\{s\}$, and their loop formulas are satisfied by a
non-stable model $\{p,q,r\}$. In fact, maximal loop $\{p,q,r\}$
is not even an elementary loop of~$\Pi_1$.
Similarly, it is not sufficient to consider maximal elementary loops
only. If we replace ``elementary loops'' in the statement of
Theorem~\ref{thm:lf}~(d) with  ``maximal elementary loops,'' then the
modified statement is incorrect. For instance, the program 
\[
\ba l
  p \ar q, \no\ p \\
  q \ar p, \no\ p \\
  p \ar\ .
\ea
\]
has two models, $\{p\}$ and $\{p,q\}$, among which the latter is not
stable. On the other hand, the only maximal elementary loop of the
program is $\{p,q\}$, and its loop formula $p\land q\rar\top$ is
satisfied by both models, so that this loop formula alone is not
sufficient to refute the stability of $\{p,q\}$. (Model $\{p,q\}$ does
not satisfy the loop formula of~$\{q\}$, which is $q\rar p\land\neg p$.)

However, in the following we show that, if we consider the
``relevant'' part of a program w.r.t. a given interpretation,
it is sufficient to restrict attention to maximal elementary loops. 
For any nondisjunctive program $\Pi$ and any set $X$ of atoms,
by $\Pi_X$, we denote the set of all rules~(\ref{abf-nd}) of~$\Pi$ 
such that $X\models B,F$. 
The following proposition states that all nonempty proper subsets of
an elementary loop of~$\Pi_X$ are externally supported by~$\Pi$
w.r.t.~$X$.

\begin{prop} \label{prop:max-elm}
For any nondisjunctive program~$\Pi$, any set~$X$ of atoms, and 
any elementary loop~$Y$ of~$\Pi_X$, $X$ satisfies ${\it ES}_\Pi(Z)$
for all nonempty proper subsets~$Z$ of~$Y$.
\end{prop}

Proposition~\ref{prop:max-elm} tells us that any elementary loop~$Y$
of~$\Pi_X$ that is unfounded by~$\Pi$ w.r.t.~$X$ is maximal among the
elementary loops of~$\Pi_X$. From this, we obtain the following result.

\setcounter{mythmc}{1}
\addtocounter{mythmc}{-1}
\begin{mythm}[e]
The following condition is equivalent to each of conditions~(a)--(c)
in Theorem~\ref{thm:lf}:
\begin{itemize}
\item[(e)]  $X$ satisfies ${\it LF}_\Pi(Y)$ for every set $Y$ of atoms 
   such that $Y$ is 
   \begin{itemize}
   \item a maximal elementary loop of~$\Pi_X$, or 
   \item a singleton whose atom occurs in~$\Pi$.
   \end{itemize} 
\end{itemize}
\end{mythm}

Given a nondisjunctive program $\Pi$ and a set $X$ of atoms, we say
that a set $Y$ of atoms that occur in $\Pi$ is {\sl elementarily
unfounded} by~$\Pi$ w.r.t.~$X$ if $Y$ is 
\begin{itemize}
\item  an elementary loop of~$\Pi_X$ that is unfounded by~$\Pi$
  w.r.t.~$X$ or 
\item  a singleton that is unfounded by $\Pi$ w.r.t.~$X$.\footnote{%
Elementarily unfounded sets are closely related to ``active elementary
loops'' defined in~\cite{gebsch05a}. We further investigate this
relationship in Section~\ref{sec:comparison}.}
\end{itemize}
Proposition~\ref{prop:max-elm} tells us that any non-singleton
elementarily unfounded set for~$\Pi$ w.r.t.~$X$ is a maximal
elementary loop of~$\Pi_X$.

It is clear from the definition that every elementarily unfounded set 
for~$\Pi$ w.r.t.~$X$ is an elementary loop of~$\Pi$ and that it is also 
unfounded by~$\Pi$ w.r.t.~$X$. However, the converse does not hold in
general. For instance, $\{p,q\}$ is an elementary loop that is
unfounded by the program
\[ 
\ba l
 p \ar q, \no\ r \\
 q \ar p, \no\ r \
\ea
\]
w.r.t.~$\{p,q,r\}$, but $\{p,q\}$ is not an elementarily unfounded set
w.r.t.~$\{p,q,r\}$.

The following corollary, which follows from
Proposition~\ref{prop:max-elm}, states that all nonempty proper
subsets of an elementarily unfounded set are externally supported. It
is essentially a reformulation of Theorem~5 from \cite{gebsch05a}.

\begin{cor} \label{cor:elm-uf}
For any nondisjunctive program~$\Pi$,  any set $X$ of atoms, and  
any elementarily unfounded set $Y$ for~$\Pi$ w.r.t.~$X$, $X$ does
not satisfy ${\it ES}_\Pi(Y)$, but satisfies ${\it ES}_\Pi(Z)$ for all
nonempty proper subsets~$Z$ of~$Y$.
\end{cor}

Corollary~\ref{cor:elm-uf} tells us that elementarily unfounded sets 
form an ``anti-chain'': one of them cannot be a proper subset of
another. (On the other hand, an elementary loop may contain another
elementary loop as its proper subset.)
Also it tells us that elementarily unfounded sets are minimal among
the nonempty unfounded sets occurring in $\Pi$. Interestingly, the
converse also holds. 

\begin{thm}\label{thm:min-uf}
For any nondisjunctive program $\Pi$ and any sets $X$, $Y$ of atoms,
$Y$ is an elementarily unfounded set for~$\Pi$ w.r.t.~$X$ iff $Y$ is
minimal among the nonempty sets of atoms occurring in~$\Pi$ that are
unfounded by~$\Pi$ w.r.t.~$X$.
\end{thm}

Notably, the correspondence between elementarily unfounded sets and
minimal nonempty unfounded sets has also led to an alternative
characterization of UE-models~\cite{gesctowo08a}, which characterizes
uniform equivalence~\cite{eitfin03a} of nondisjunctive programs as
well as disjunctive programs.

Similar to Theorem~\ref{thm:lf}~(b$'$), Theorem~\ref{thm:lf}~(e) can
be stated in terms of elementarily unfounded sets, thereby restricting
attention to minimal nonempty unfounded sets.

\setcounter{mythmc}{1}
\addtocounter{mythmc}{-1}
\begin{mythm}[e$'$]
The following condition is equivalent to each of conditions~(a)--(c)
in Theorem~\ref{thm:lf}:
\begin{itemize}
\item[(e$'$)]  $X$ contains no elementarily unfounded sets 
               for~$\Pi$ w.r.t.~$X$.
\end{itemize}
\end{mythm}

The notion of an elementarily unfounded set may help improve
computation performed by SAT-based answer set solvers. Since there are
exponentially many ``relevant'' loops in the worst
case~\cite{lifraz04a}, SAT-based answer set solvers do not add all
loop formulas at once. Instead, they check whether a model returned by
a SAT solver is stable. If not, a loop formula that is not satisfied
by the model is added, and the SAT solver is invoked again. This
process is repeated until a stable model is found or the search space
is exhausted. 
In view of Theorem~\ref{thm:lf}~(e$'$), it is sufficient to restrict
attention to elementarily unfounded sets during the computation.
This ensures that the considered loop formulas belong to elementary
loops. Since every elementary loop is a loop, but not vice versa, the
computation may involve fewer loop formulas overall than in the case
when arbitrary loops are considered. However, whether this idea will
lead to more efficient computation in practice requires further
investigation.


\subsection{Recognizing Elementary Loops of Nondisjunctive
  Programs}\label{sec:dec-nd} 

The definition of an elementary loop given in Section~\ref{ssec:elm}
involves considering all its nonempty proper subsets (or at least all
elementary loops that are subsets). This may seem to imply that
deciding whether a given set of atoms is an elementary loop is a
computationally hard problem. However, Gebser and
Schaub~\citeyear{gebsch05a} showed that this is not the case for
nondisjunctive programs. They also noted that the notion of a positive
dependency graph is not expressive enough to distinguish between
elementary and non-elementary loops (Program~$\Pi_1$ and the program
in~(\ref{ex2}) have the same dependency graph, but their elementary
loops are different), and instead used the rather complicated notion
of a {\em body-head dependency graph} to identify elementary loops. 
In this section, we simplify their result by still referring to
a positive dependency graph. We show that removing some
{\em unnecessary} edges from a positive dependency graph is just enough
to distinguish between elementary and non-elementary loops.

\begin{figure}
\begin{center}
 \begin{picture}(120,10)(0,3)
 \put(0,0){$p$}
 \put(90,0){$q$}
 \put(45.5,0){$r$}

 \qbezier(8,5)(25,11)(42,5) \put(43,4.6){\vector(3,-1){1}}

 \qbezier(88,5)(71,11)(54,5) \put(53,4.6){\vector(-3,-1){1}}
 \end{picture}
\caption{The elementary subgraph of $\{p,q,r\}$ for Program $\Pi_1$}
\label{fig:dep2}
\end{center}
\end{figure}
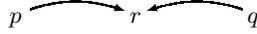

For any set $X$ of atoms that occur in a nondisjunctive program~$\Pi$,
we define:
\begin{eqnarray*} 
{\it EC}_\Pi^0(X) &=& \emptyset \ ,\\
{\it EC}_\Pi^{i+1}(X) &=& 
   \{ (a_1,b) \mid 
\begin{array}[t]{@{}l@{}}
\text{there is a rule~(\ref{abf-nd}) in $\Pi$ such that $a_1\in X$,} \\
\text{$b\in B\cap X$, and all atoms in $B\cap X$ belong to the} \\
\text{same strongly connected component in $(X,{\it EC}^i_\Pi(X))$}\} \ , 
\end{array}
\\
{\it EC}_\Pi(X) &=& \mbox{$\bigcup$}_{i\ge 0}
{\it EC}_\Pi^i(X) \ . 
\end{eqnarray*}
This is a ``bottom-up'' construction based on strongly connected
components, i.e., maximal strongly connected subgraphs of a given
graph. Thus ${\it EC}_\Pi^i(X)$ is a subset of ${\it EC}_\Pi^{i+1}(X)$,
and the graph $(X,{\it EC}_\Pi(X))$ is a subgraph of the positive
dependency graph of~$\Pi$. We call the graph $(X,{\it EC}_\Pi(X))$ the
{\sl elementary subgraph} of $X$ for $\Pi$. 
Figure~\ref{fig:dep2} shows the elementary subgraph of $\{p,q,r\}$ 
for~$\Pi_1$, which is not strongly connected.

The following theorem is similar to Theorem~10 from \cite{gebsch05a},
but instead of referring to the notion of a body-head dependency
graph, it refers to the notion of an elementary subgraph. 

\begin{thm} \label{thm:ec-tr}
For any nondisjunctive program $\Pi$ and any nonempty set~$X$ of atoms
that occur in~$\Pi$, $X$ is an elementary loop of~$\Pi$ iff the
elementary subgraph of $X$ for~$\Pi$ is strongly connected.
\end{thm}

Since an elementary subgraph can be constructed in polynomial time,
the problem of deciding whether a given set of atoms is an elementary
loop of a nondisjunctive program is tractable.

\subsection{Comparison with Gebser-Schaub Definition} 
\label{sec:comparison}

In this section, we compare our reformulation of elementary loops with
the original definition by Gebser and Schaub~\citeyear{gebsch05a} for
nondisjunctive programs. 

Let $\Pi$ be a nondisjunctive program. A loop of $\Pi$ is called {\sl
  trivial} if it consists of a single atom such that the
dependency graph of $\Pi$ does not contain an edge from the atom 
to itself, and {\sl nontrivial} otherwise.\footnote{%
In~\cite{lin04} and \cite{gebsch05a}, loops were defined to be
nontrivial.}
For a nontrivial loop~$L$ of~$\Pi$, let
\begin{itemize}
\item  $R^-_\Pi(L) = \{(\ref{abf-nd}) \in\Pi 
        \mid a_1\in L,\ B\cap L=\emptyset \}$, and 
\item  $R^+_\Pi(L) = \{(\ref{abf-nd}) \in\Pi 
        \mid a_1\in L,\ B\cap L\ne\emptyset \}$.
\end{itemize}

\begin{definition} [\cite{gebsch05a}]\label{def:elmloop}
A nontrivial loop $L$ of a nondisjunctive program $\Pi$ is called a
{\sl GS-elementary loop} of~$\Pi$ if 
$R^-_\Pi(L')\cap R^+_\Pi(L)\ne\emptyset$ 
for all proper subsets~$L'$ of~$L$ that are nontrivial loops of~$\Pi$.
\end{definition}

\begin{prop} \label{prop:gs}
For any nondisjunctive program $\Pi$ and any set $L$ of atoms, $L$ is
a GS-elementary loop of~$\Pi$ iff $L$ is a nontrivial elementary loop
of~$\Pi$.
\end{prop}

\proof
{\sl From left to right:} 
Assume that $L$ is a GS-elementary loop of~$\Pi$. If~$L$ is a
singleton, it is a (nontrivial) elementary loop according to our
definition. If~$L$ is not a singleton, we have that 
$R^-_\Pi(L')\cap R^+_\Pi(L)\ne\emptyset$ for any proper subset~$L'$
of~$L$ that is a nontrivial loop of~$\Pi$.
In other words, there is a rule~(\ref{abf-nd}) in~$\Pi$ such that
\beq
a_1\in L' ,
\eeq{gs1}
\beq
B\cap L'= \emptyset ,
\eeq{gs3}
and
\beq
B\cap (L\setminus L')\ne\emptyset .
\eeq{gs2}
We thus have that~$L'$ is outbound in~$L$ for~$\Pi$.
Furthermore, for any trivial loop~$\{a_1\}$ of~$\Pi$ contained in~$L$,
there must be a rule~(\ref{abf-nd}) in~$\Pi$ such that 
$B\cap (L\setminus\{a_1\})\ne\emptyset$, as $L$ cannot be a loop of
$\Pi$ otherwise.
Since $\{a_1\}$ is trivial, $B\cap\{a_1\}=\emptyset$,
so that $\{a_1\}$ is outbound in~$L$ for~$\Pi$.
By Proposition~\ref{prop:elm-alt},
it follows that~$L$ is a (nontrivial) elementary loop of~$\Pi$.

\medskip\noindent
{\sl From right to left:}
Assume that $L$ is a nontrivial elementary loop of~$\Pi$. From the
definition of an elementary loop, it follows that any proper
subset~$L'$ of~$L$ that is a nontrivial loop of~$\Pi$
is outbound in~$L$ for~$\Pi$. That is, there is a rule~(\ref{abf-nd})
in~$\Pi$ such that (\ref{gs1}), (\ref{gs3}), and~(\ref{gs2}) hold, so
that $L$ is a GS-elementary loop of~$\Pi$.
\qed

For a nondisjunctive program~$\Pi$ and a set~$X$ of atoms, a loop~$L$
of~$\Pi_X$ is a GS-elementary loop of~$\Pi_X$ iff $L$ is a nontrivial
elementary loop of~$\Pi_X$. Thus  an {\em active elementary loop}
of~$\Pi$ according to~\cite{gebsch05a} is a nontrivial elementary loop
of~$\Pi_X$ that is unfounded by~$\Pi$ w.r.t.~$X$. Hence, any active
elementary loop~$L$ of~$\Pi$ is an elementarily unfounded set
for~$\Pi$ w.r.t.~$X$, while the converse does not hold in general
if~$L$ is a singleton.

In fact, there are a few differences between
Definition~\ref{def:elmloop} and our definition of an elementary
loop. First, our definition of an elementary loop does not a priori
assume that its atoms form a loop. Rather, the fact that an elementary
loop is a loop follows from its definition in view of
Proposition~\ref{prop:loop-alternative}.
Second, the two definitions do not agree on trivial loops: a trivial
loop is an elementary loop, but not a GS-elementary loop. This
originates from the difference between the definition of a loop
in~\cite{lin04} and its reformulation given in~\cite{lee05}.
As shown in the main theorem from \cite{lee05}, identifying a trivial
loop as a loop admits a simpler reformulation of the Lin-Zhao theorem 
by allowing us to view completion formulas~\cite{cla78} as a special
case of loop formulas. 
Furthermore, the reformulated definition of an elementary loop enables
us to identify a close relationship between maximal elementary loops
(elementarily unfounded sets) and minimal nonempty unfounded sets. 

Importantly, trivial loops allow us to extend the notion of an
elementary loop to disjunctive programs without producing
counterintuitive results. For instance, consider the following
disjunctive program:
\beq
\ba l
 p\ ;\ q \ar r \\  
 p\ ;\ r \ar q \\
 q\ ;\ r \ar p \ . 
\ea
\eeq{ex-uf}
The nontrivial loops of this program are $\{p,q\}$, $\{p,r\}$, $\{q,r\}$,
and $\{p, q, r\}$, but not the singletons $\{p\}$, $\{q\}$, and $\{r\}$.
If we were to extend GS-elementary loops to disjunctive programs, 
a natural extension would say that $\{p,q,r\}$ is a GS-elementary 
loop since $\{p,q\}$, $\{p,r\}$, and $\{q,r\}$ are ``outbound''
in~$\{p,q,r\}$. But note that $\{p,q,r\}$ is unfounded
w.r.t.~$\{p,q,r\}$; moreover, every singleton is also unfounded
w.r.t~$\{p,q,r\}$. This is in contrast with
Proposition~\ref{prop:max-elm}, according to which all nonempty 
proper subsets of an elementary loop should be externally supported. 
The next section shows that such an anomaly does not arise with our
definition of an elementary loop that is extended to disjunctive
programs. 


\section{Disjunctive Programs}\label{sec:elm-dis}

After providing the relevant background on disjunctive programs, this
section generalizes the notions of an elementary loop and an
elementarily unfounded set to disjunctive programs. We also provide 
the proofs of the generalizations of the statements given in the
previous section; such generalized results also apply to the class of
nondisjunctive programs as a fragment of disjunctive
programs. Furthermore, we show that, in contrast to nondisjunctive
programs, recognizing an elementary loop is intractable in the case of
arbitrary disjunctive programs, but stays tractable under a certain
syntactic condition.


\subsection{Background}

A {\sl disjunctive rule} is an expression of the form   
\beq
\ba l
  a_1;\dots; a_k \ar  
         a_{k+1},\dots, a_l, 
         \no\ a_{l+1},\dots, \no\ a_m, 
         \no\ \no\ a_{m+1},\dots, \no\ \no\ a_n
\ea
\eeq{dis}
where $n\ge m\ge l\ge k\ge 0$ and $a_1,\dots, a_n$ are propositional
atoms.  A {\sl disjunctive program} is a finite set of disjunctive
rules. Note that any program with nested expressions can be turned
into an equivalent program whose rules are of the
form~(\ref{dis})~\cite{lif99d}. 

We will identify a disjunctive rule~(\ref{dis}) with the propositional 
formula  
\[
\ba l
  a_{k+1}\land\cdots\land a_l\land\neg a_{l+1}\land\cdots\land\neg a_m \land 
  \neg\neg a_{m+1}\land\cdots\land\neg\neg a_n 
  \, \rar\, a_1\lor\cdots\lor a_k 
\ea
\]
and will often write~(\ref{dis}) as  
\beq
  A\, \ar\, B, F
\eeq{abf-dis}
where $A$ is $a_1,\dots,a_k$, $B$ is $a_{k+1},\dots,a_l$, and $F$ is  
\[
   \no\ a_{l+1},\dots,\no\ a_m,\no\ \no\ a_{m+1},\dots,\no\ \no\ a_n\ .
\]
We will sometimes identify $A$ and $B$ with their corresponding sets of atoms.

The reduct $\Pi^X$ of a disjunctive program $\Pi$ w.r.t. a set~$X$ of
atoms is obtained from~$\Pi$ by deleting each rule~(\ref{abf-dis})
such that $X\not\models F$, and replacing each remaining
rule~(\ref{abf-dis}) with $A\ar B$. A set~$X$ of atoms is a {\em
  stable model}, also called an {\em answer set}, of~$\Pi$ if~$X$ is
minimal among the sets of atoms that satisfy~$\Pi^X$.

The definition of a {\sl (positive) dependency graph} is extended to a
disjunctive program~$\Pi$ in the straightforward way: the vertices of
the graph are the atoms occurring in~$\Pi$, and its edges go from the
elements of~$A$ to the elements of~$B$ for all rules (\ref{abf-dis})
of~$\Pi$. With this extended definition of a dependency graph, the
definition of a loop for a nondisjunctive program is
straightforwardly extended to a disjunctive program. 

For any set $Y$ of atoms, the {\sl external support formula} of~$Y$
for a disjunctive program~$\Pi$, denoted by~${\it ES}_\Pi(Y)$, is 
the disjunction of conjunctions  
\[ 
   B\land F\land \bigwedge_{a\in A\setminus Y} \neg a 
\]
for all rules~(\ref{abf-dis}) of~$\Pi$ such that $A\cap Y\ne\emptyset$
and $B\cap Y = \emptyset$.
When $\Pi$ is a nondisjunctive program, this definition reduces to the
definition of ${\it ES}_\Pi(Y)$ for nondisjunctive programs given earlier.
As before, we say that $Y$ is {\em externally supported} by $\Pi$
w.r.t. a set $X$ of atoms if $X\models{\it ES}_\Pi(Y)$; $Y$ is {\em
  unfounded} $\Pi$ w.r.t.~$X$ if $X\not\models{\it ES}_\Pi(Y)$. 

The notion of ${\it LF}_\Pi(Y)$ and the term {\sl (conjunctive) loop
  formula} similarly apply to formulas~(\ref{lf}) when $\Pi$ is a
disjunctive program. 

As shown in \cite{lee05}, Theorem~\ref{thm:lf} remains correct  after
replacing ``nondisjunctive program'' in its statement  with
``disjunctive program.''

\setcounter{mythmc}{1}
\addtocounter{mythmc}{-1}
\begin{mythmd} [\cite{lee05}]
For any disjunctive program~$\Pi$ and any set~$X$ of atoms that occur
in~$\Pi$, if $X$ is a model of~$\Pi$, then the following conditions
are equivalent: \footnote{%
Superscript $^{\rm d}$ indicates that the statement is a generalization to
disjunctive programs.} 
\begin{itemize}
\item[(a)]  $X$ is a stable model of~$\Pi$;
\item[(b)]  $X$ satisfies~${\it LF}_\Pi(Y)$ for all nonempty sets $Y$ of
  atoms that occur in~$\Pi$;            
\item[(b$'$)] $X$ contains no nonempty unfounded sets for~$\Pi$
  w.r.t.~$X$;
\item[(c)]  $X$ satisfies~${\it LF}_\Pi(Y)$ for all loops $Y$ of~$\Pi$.
\end{itemize}
\end{mythmd}

For instance, the loop formulas of the seven loops of the program
in~(\ref{ex-uf}) are:
\beq
\ba l
   p \rar (r\land \neg q) \lor (q\land \neg r) \\
   q \rar (r\land \neg p) \lor (p\land \neg r) \\
   r \rar (q\land \neg p) \lor (p\land \neg q) \\
   p \land q \rar r \\
   p \land r \rar q  \\
   q \land r \rar p \\
   p \land q \land r \rar \bot \ . 
\ea
\eeq{ex-dlf} 
$\emptyset$ is the only model of~(\ref{ex-dlf}) and it is the only
stable model of (\ref{ex-uf}) in accordance with the equivalence
between (a) and (c) in Theorem$^{\rm d}$~\ref{thm:lf}. 

\subsection{Elementary Loops of Disjunctive Programs}

In this section, we generalize the definition of an elementary loop to
disjunctive programs. 

A loop of a disjunctive program can be defined without referring to a
dependency graph, as in Proposition~\ref{prop:loop-alternative}.

\setcounter{mythmc}{1}
\addtocounter{mythmc}{-1}
\begin{mypropd}
For any disjunctive program~$\Pi$ and any nonempty set~$X$ of atoms
that occur in~$\Pi$, $X$ is a loop of~$\Pi$ iff, for every nonempty
proper subset~$Y$ of $X$, there is a rule~(\ref{abf-dis}) in~$\Pi$
such that $A\cap Y\ne\emptyset$ and $B\cap (X\setminus
Y)\ne\emptyset$.
\end{mypropd}

\proof
{\sl From left to right:} Assume that $X$ is a loop of $\Pi$. If $X$ is 
a singleton, it is clear. If $X$ is not a singleton, take any nonempty
proper subset~$Y$ of $X$. Since both $Y$ and $X\setminus Y$ are 
nonempty, there is a path from some atom in~$Y$ to some atom in
$X\setminus Y$ in the dependency graph of~$\Pi$ such that all
vertices in the path belong to~$X$.
This implies that there is an edge from an atom in~$Y$ to an atom in
$X\setminus Y$, i.e., $A\cap Y\ne\emptyset$ and $B\cap (X\setminus
Y)\ne\emptyset$ for some rule~(\ref{abf-dis}) in~$\Pi$. 

\medskip\noindent
{\sl From right to left:}
Assume that $X$ is not a loop of $\Pi$. Then the subgraph of the
dependency graph of~$\Pi$ induced by~$X$ is not strongly
connected. Consequently, there is a nonempty proper subset~$Y$ of~$X$
such that no edge connects an atom in~$Y$ to an atom in~$X\setminus
Y$.  This implies that there is no rule~(\ref{abf-dis}) in~$\Pi$ such
that $A\cap Y\ne\emptyset$ and $B\cap (X\setminus Y)\ne\emptyset$.
\qed

For any set $X$ of atoms and any subset~$Y$ of~$X$, we say that~$Y$ is
{\sl outbound} in~$X$ for a disjunctive program~$\Pi$ if there is a
rule~(\ref{abf-dis}) in~$\Pi$ such that 
\begin{itemize}
\item $A\cap Y\ne\emptyset$,
\item $B\cap (X\setminus Y)\ne\emptyset$, 
\item $A\cap (X\setminus Y)=\emptyset$, and 
\item $B\cap Y=\emptyset$.
\end{itemize}
As with nondisjunctive programs, for any nonempty set~$X$ of atoms
that occur in~$\Pi$, we say that $X$ is an {\sl elementary loop}
of~$\Pi$ if all nonempty proper subsets of~$X$ are outbound in~$X$
for~$\Pi$. Clearly, every singleton whose atom occurs in~$\Pi$ is an
elementary loop of~$\Pi$, and every elementary loop of~$\Pi$ is a loop
of~$\Pi$. The definition of an elementary loop of a disjunctive
program is stronger than the alternative characterization of a loop
provided in Proposition$^{\rm d}$~\ref{prop:loop-alternative}: it
requires that the rule~(\ref{abf-dis}) satisfy two additional
conditions, $A \cap (X\setminus Y) = \emptyset$
and $B \cap Y = \emptyset$.

To illustrate the definition of an elementary loop of a disjunctive
program, consider the loop $\{p,q,r\}$ of the program
in~(\ref{ex-uf}). The loop is not an elementary loop because, for
instance, $\{p\}$ is not outbound in~$\{p,q,r\}$: although the first
two rules~(\ref{abf-dis}) in~(\ref{ex-uf}) are such that $A\cap
\{p\}\ne\emptyset$, $B\cap\{q,r\}\ne\emptyset$, and $B\cap
\{p\}=\emptyset$, we also have that $A\cap\{q,r\}\ne\emptyset$ for
each of them. Similarly, $\{q\}$ and $\{r\}$ are not outbound
in~$\{p,q,r\}$. On the other hand, the remaining loops of the program,
$\{p\}$, $\{q\}$, $\{r\}$, $\{p,q\}$, $\{p,r\}$, and $\{q,r\}$,
are elementary loops.

With the extended definitions given above,
Propositions~\ref{prop:es},~\ref{prop:elm-alt},~\ref{prop:elm-lf-entail}
and Theorem~\ref{thm:lf}~(d) remain correct after replacing
``nondisjunctive program'' in their statements with ``disjunctive
program.''   In the following, we present proofs for these
generalizations.

\setcounter{mythmc}{2}
\addtocounter{mythmc}{-1}
\begin{mypropd}
For any  disjunctive program $\Pi$ and any sets $X$, $Y$, $Z$ of
atoms such that $Z\subseteq Y\subseteq X$, if $Z$ is not outbound in
$Y$ for~$\Pi$ and $X\models{\it ES}_\Pi(Z)$, then $X\models{\it ES}_\Pi(Y)$.
\end{mypropd}

\proof
Assume that $Z$ is not outbound in $Y$ for~$\Pi$ and that
$X\models{\it ES}_\Pi(Z)$. From the latter, it follows that there is a
rule~(\ref{abf-dis}) in~$\Pi$ such that
\beq
  A\cap Z\ne\emptyset\ ,
\eeq{az1}
\beq
  B\cap Z=\emptyset\ ,
\eeq{bz1}
\beq
  X\models B, F\ ,
\eeq{xbf1}
and
\beq
  X\cap (A\setminus Z)=\emptyset\ .
\eeq{xaz1}
 From~(\ref{az1}), since $Z\subseteq Y$,
\beq
  A\cap Y\ne\emptyset\ .
\eeq{ay1}
 From~(\ref{xaz1}), since $Z\subseteq Y\subseteq X$, 
\beq
  X\cap (A\setminus Y)=\emptyset
\eeq{xay1.1}
and
$$
  Y\cap(A\setminus Z) = \emptyset\ ,
$$
where the latter is equivalent to
\beq
  A\cap (Y\setminus Z) = \emptyset\ .
\eeq{ay1z1}
Since $Z$ is not outbound in $Y$ for~$\Pi$, from~(\ref{az1}),
(\ref{bz1}), and (\ref{ay1z1}), it follows that
$$ 
 B\cap (Y\setminus Z) = \emptyset\ ,
$$ 
which, in combination with~(\ref{bz1}), gives us that
\beq
 B\cap Y=\emptyset\ .
\eeq{by1}
Finally, from~(\ref{xbf1}), (\ref{ay1}), (\ref{xay1.1}), and
(\ref{by1}), we conclude that $X\models{\it ES}_\Pi(Y)$.
\qed

\setcounter{mythmc}{3}
\addtocounter{mythmc}{-1}
\begin{mypropd}
For any disjunctive program~$\Pi$ and any nonempty set $X$ of atoms
that occur in~$\Pi$, $X$ is an elementary loop of~$\Pi$ iff all
proper subsets of $X$ that are elementary loops of~$\Pi$  are outbound
in $X$ for $\Pi$.
\end{mypropd}

\proof
From left to right is clear.

\medskip\noindent
{\sl From right to left:}
Assume that $X$ is not an elementary loop of~$\Pi$. Then there is a
nonempty proper subset~$Y$ of~$X$ that is not outbound in~$X$
for~$\Pi$.
If $Y$ is an elementary loop of~$\Pi$, it is clear. 
Otherwise, there is a nonempty proper subset~$Z$ of~$Y$ that is not
outbound in~$Y$ for~$\Pi$. For the sake of contradiction,
assume that $Z$ is outbound in~$X$ for~$\Pi$, i.e.,
that there is a rule~(\ref{abf-dis}) in~$\Pi$ such that
\beq  A\cap Z\ne\emptyset\ , \eeq{s1}
\beq  B\cap (X\setminus Z)\ne\emptyset\ , \eeq{s2}
\beq  A\cap (X\setminus Z)= \emptyset\ , \eeq{s3}
and
\beq  B\cap Z = \emptyset\ . \eeq{s4} 
From~(\ref{s1}) and~(\ref{s3}), since $Z\subseteq Y\subseteq X$,
\beq
  A\cap Y\ne\emptyset\ ,
\eeq{s5}
\beq
  A\cap (X\setminus Y)= \emptyset\ ,
\eeq{s6}
and
\beq  
  A\cap (Y\setminus Z)= \emptyset\ .
\eeq{s7}
Since~$Z$ is not outbound in~$Y$ for~$\Pi$, from (\ref{s1}),
(\ref{s4}), and (\ref{s7}), it follows that
\beq\nonumber  
  B\cap (Y\setminus Z)=\emptyset\ ,
\eeq{s8}
which, in combination with~(\ref{s2}) and (\ref{s4}), gives us that
\beq  
  B\cap Y=\emptyset 
\eeq{s9}
and
\beq  
  B\cap (X\setminus Y)\ne\emptyset\ . 
\eeq{s10}
However, (\ref{s5}), (\ref{s6}), (\ref{s9}), and (\ref{s10}) together
contradict that $Y$ is not outbound in~$X$ for~$\Pi$, from which we
conclude that $Z$ is not outbound in~$X$ for~$\Pi$.
We have thus shown that every nonempty proper subset of~$X$ that is
not outbound in~$X$ for~$\Pi$ and not an elementary loop of~$\Pi$
contains in turn a nonempty proper subset that is not outbound in~$X$
for~$\Pi$. Since $X$ is finite, there is some (not necessarily unique)
minimal nonempty proper subset of~$X$ that is not outbound in~$X$ for~$\Pi$,
and such a subset must be an elementary loop of~$\Pi$.
\qed

\setcounter{mythmc}{4}
\addtocounter{mythmc}{-1}
\begin{mypropd}
For any disjunctive program $\Pi$ and any nonempty set $Y$ of
  atoms that occur in~$\Pi$, there is an elementary loop $Z$ of $\Pi$
  such that $Z$ is a subset of $Y$ and ${\it LF}_\Pi(Z)$ entails
  ${\it LF}_\Pi(Y)$.
\end{mypropd}

\proof
If $Y$ is an elementary loop of~$\Pi$, it is clear.
Otherwise, by Proposition$^{\rm d}$~\ref{prop:elm-alt}, some proper
subset~$Z$ of~$Y$ is an elementary loop of~$\Pi$ that is not outbound
in~$Y$ for~$\Pi$. Take any set $X$ of atoms such that
$X\models{\it LF}_\Pi(Z)$. If $Y\not\subseteq X$,
then $X\not\models\bigwedge_{a\in Y} a$ and $X\models{\it LF}_\Pi(Y)$.
If $Y\subseteq X$, $X\models\bigwedge_{a\in Z} a$ and
$X\models{\it ES}_\Pi(Z)$, and, by Proposition$^{\rm d}$~\ref{prop:es},
we conclude that $X\models{\it ES}_\Pi(Y)$ and $X\models{\it LF}_\Pi(Y)$.
\qed

\setcounter{mythmc}{1}
\addtocounter{mythmc}{-1}
\begin{mythmd}[d]
The following condition is equivalent to each of conditions~(a)--(c)
in Theorem$^{\rm d}$~\ref{thm:lf}:
\begin{itemize}
\item[(d)]  $X$ satisfies~${\it LF}_\Pi(Y)$ for all elementary loops~$Y$
  of~$\Pi$. 
\end{itemize}
\end{mythmd}

\proof
We show the equivalence between (b) and (d). From (b) to (d) is clear,
and from (d) to (b) follows immediately from Proposition$^{\rm
  d}$~\ref{prop:elm-lf-entail}.
\qed

For instance, for the program in~(\ref{ex-uf}), the loop formula of
non-elementary loop $\{p,q,r\}$ (the last one in~(\ref{ex-dlf})) can
be disregarded in view of Theorem$^{\rm d}$~\ref{thm:lf}~(d).


\subsection{Elementarily Unfounded Sets for Disjunctive
  Programs}\label{sec:eusdp} 

Let $\Pi$ be a disjunctive program. For any sets $X$, $Y$ of atoms,
by~$\Pi_{X,Y}$ we denote the set of all rules~(\ref{abf-dis}) of~$\Pi$
such that $X\models B,F$ and $X\cap (A\setminus Y)=\emptyset$. 
That is, the program~$\Pi_{X,Y}$ contains all rules of~$\Pi$ that can
provide supports for~$Y$ w.r.t.~$X$.
If $Y=X$, we also denote~$\Pi_{X,Y}$ by~$\Pi_X$.
When $\Pi$ is a nondisjunctive program, this definition reduces to 
the definition of~$\Pi_X$ for nondisjunctive programs given earlier.
Furthermore, when~$\Pi$ is nondisjunctive and $Y$ is not a singleton,
then $Y$ is an elementary loop of~$\Pi_{X,Y}$ iff $Y$ is an elementary
loop of~$\Pi_X$.

We extend the definition of an elementarily unfounded set to
disjunctive programs by replacing ``$\Pi_X$'' with ``$\Pi_{X,Y}$'':
for a disjunctive program~$\Pi$ and a set~$X$ of atoms,
we say that a set~$Y$ of atoms that occur in~$\Pi$ is {\em
  elementarily unfounded} by~$\Pi$ w.r.t.~$X$ if $Y$ is 
\begin{itemize}
\item  an elementary loop of~$\Pi_{X,Y}$ that is unfounded by~$\Pi$
  w.r.t.~$X$ or 
\item  a singleton that is unfounded by $\Pi$ w.r.t.~$X$.
\end{itemize}
It is clear from the definition that every elementarily unfounded set
for~$\Pi$ w.r.t.~$X$ is an elementary loop of~$\Pi$ and that it is
also unfounded by~$\Pi$ w.r.t.~$X$.

For instance, let~$\Pi$ be the program (\ref{ex-uf}).
The program $\Pi_{\{p,q,r\},\{p,q\}}$ consists of the first rule
in~(\ref{ex-uf}), so that $\{p,q\}$ is not an elementary loop of
$\Pi_{\{p,q,r\},\{p,q\}}$. On the other hand, $\Pi_{\{p,q\},\{p,q\}}$
consists of the last two rules in~(\ref{ex-uf}), and $\{p,q\}$ is an
elementary loop of $\Pi_{\{p,q\},\{p,q\}}$.
Since $\{p,q\}$ is also unfounded by~$\Pi$ w.r.t.\ $\{p,q\}$,
it is an elementarily unfounded set for~$\Pi$ w.r.t.\ $\{p,q\}$.

Proposition~\ref{prop:max-elm}, Corollary~\ref{cor:elm-uf},
and Theorem~\ref{thm:min-uf} remain correct after replacing
``nondisjunctive program'' in their statements with ``disjunctive
program,'' and ``$\Pi_X$'' with ``$\Pi_{X,Y}$.''

\setcounter{mythmc}{5}
\addtocounter{mythmc}{-1}
\begin{mypropd}
For any disjunctive program~$\Pi$, any set $X$ of atoms, and 
any elementary loop $Y$ of $\Pi_{X,Y}$, $X$ satisfies ${\it ES}_\Pi(Z)$
for all nonempty proper subsets~$Z$ of $Y$.
\end{mypropd}

\proof
From the fact that $Y$ is an elementary loop of $\Pi_{X,Y}$, it
follows that any nonempty proper subset $Z$ of $Y$ is outbound in~$Y$
for $\Pi_{X,Y}$.
If $Y$ is not a singleton, this implies that~$Y$ is a subset of~$X$
and that, for each nonempty proper subset~$Z$ of~$Y$, there is a
rule~(\ref{abf-dis}) in~$\Pi$ such that 
\beq 
  A\cap Z\ne\emptyset\ ,
\eeq{m-az} 
\beq
  A\cap (Y\setminus Z)=\emptyset\ ,
\eeq{m-ayz}
\beq 
  B\cap Z=\emptyset\ ,
\eeq{m-bz}
\beq
  X\models B,F\ , 
\eeq{m-xbf}
and 
\beq 
   X\cap (A\setminus Y)=\emptyset\ . 
\eeq{m-xay}
 From~(\ref{m-ayz}) and (\ref{m-xay}), it follows that 
\beq
  X\cap (A\setminus Z)=\emptyset\ .
\eeq{m-xaz}
Finally, from~(\ref{m-az}), (\ref{m-bz}), (\ref{m-xbf}), and
(\ref{m-xaz}), we conclude that $X\models{\it ES}_\Pi(Z)$.
\qed

\setcounter{mythmc}{1}
\addtocounter{mythmc}{-1}
\begin{mycold}
For any disjunctive program  $\Pi$,  any set $X$ of atoms, and  any
elementarily unfounded set $Y$ for~$\Pi$ w.r.t.~$X$, $X$ does not
satisfy ${\it ES}_\Pi(Y)$, but satisfies ${\it ES}_\Pi(Z)$ for all
nonempty proper subsets~$Z$ of~$Y$.
\end{mycold}

\proof
From the definition of an elementarily unfounded set,
$X\not\models{\it ES}_\Pi(Y)$, and,
by Proposition$^{\rm d}$~\ref{prop:max-elm}, 
$X\models{\it ES}_\Pi(Z)$
for all nonempty proper subsets~$Z$ of~$Y$.
\qed

\setcounter{mythmc}{2}
\addtocounter{mythmc}{-1}
\begin{mythmd}
For any disjunctive program $\Pi$ and any sets $X$, $Y$ of atoms, 
$Y$ is an elementarily unfounded set for $\Pi$ w.r.t.~$X$ iff 
$Y$ is minimal among the nonempty sets of atoms occurring in $\Pi$
that are unfounded by $\Pi$ w.r.t.~$X$.
\end{mythmd}

\proof
From left to right follows immediately from 
Corollary$^{\rm d}$~\ref{cor:elm-uf}.

\medskip\noindent
{\sl From right to left: }
Assume that $Y$ is minimal among the nonempty unfounded sets for~$\Pi$
w.r.t.~$X$ whose atoms occur in $\Pi$. If $Y$ is a singleton, it 
is elementarily unfounded by~$\Pi$ w.r.t.~$X$.
Otherwise, if $Y\not\subseteq X$, 
there is an atom $a\in (Y\setminus X)$, and one can check that
$(Y\setminus \{a\})$ is also unfounded by $\Pi$ w.r.t.~$X$, which
contradicts that $Y$ is a minimal nonempty unfounded set for~$\Pi$
w.r.t.~$X$.
Hence, from the minimality assumption on $Y$, it follows that $Y$ is a
subset of~$X$. 
It also holds that $X\models{\it ES}_\Pi(Z)$ for every nonempty proper
subset~$Z$ of~$Y$, so that there is  a rule~(\ref{abf-dis}) in $\Pi$
such that 
\beq  
   A\cap Z\ne\emptyset\ , 
\eeq{u-az}
\beq  
   B\cap Z=\emptyset\ , 
\eeq{u-bz}
\beq  
   X\models B, F\ ,
\eeq{u-xbf} 
and
\beq  
   X\cap (A\setminus Z)=\emptyset\ .
\eeq{u-xaz}
From~(\ref{u-xaz}), since $Z\subseteq Y\subseteq X$,
\beq
   A\cap (Y\setminus Z)=\emptyset
\eeq{u-ayz}
and
\beq  
   X\cap (A\setminus Y)=\emptyset\ .
\eeq{u-xay}
Since $Y$ is unfounded by $\Pi$ w.r.t.~$X$, from~(\ref{u-az}),
(\ref{u-xbf}), and~(\ref{u-xay}), it follows that 
$$ 
   B\cap Y\ne\emptyset\ ,
$$ 
which, in combination with~(\ref{u-bz}), gives us that
\beq
   B\cap (Y\setminus Z)\ne\emptyset\ .
\eeq{u-byz}
In view of~(\ref{u-xbf}) and (\ref{u-xay}), we have that the
rule~(\ref{abf-dis}) is contained in $\Pi_{X,Y}$.
 From~(\ref{u-az}), (\ref{u-bz}), (\ref{u-ayz}), and (\ref{u-byz}), 
we further conclude that $Z$ is outbound in~$Y$ for $\Pi_{X,Y}$.
Consequently, $Y$ is an elementary loop of~$\Pi_{X,Y}$ and
elementarily unfounded by $\Pi$ w.r.t.~$X$.
\qed

Theorem~\ref{thm:lf}~(e) and (e$'$) can now be extended to disjunctive
programs as follows.

\setcounter{mythmc}{1}
\addtocounter{mythmc}{-1}
\begin{mythmd}[e$'$] 
The following conditions are equivalent to each of conditions~(a)--(c)
in Theorem$^{\rm d}$~\ref{thm:lf}:
\begin{itemize}
\item[(e)]  $X$ satisfies ${\it LF}_\Pi(Y)$ for every set $Y$ of atoms
  such that $Y$ is
  \begin{itemize}
  \item  maximal among all sets $Z$ of atoms that are elementary loops
    of $\Pi_{X,Z}$ or
  \item  a singleton whose atom occurs in~$\Pi$;
  \end{itemize}
\item[(e$'$)]  $X$ contains no elementarily unfounded sets for~$\Pi$
  w.r.t.~$X$.
\end{itemize}
\end{mythmd}

\proof
We first show the equivalence between (b$'$) and (e$'$): from (b$'$)
to (e$'$) is clear, and from (e$'$) to (b$'$) follows immediately from
Theorem$^{\rm d}$~\ref{thm:min-uf}. 
Moreover, the equivalence between (e$'$) and (e) holds in view of
Proposition$^{\rm d}$~\ref{prop:max-elm}, which tells us that an
elementarily unfounded set~$Y$ for~$\Pi$ w.r.t.~$X$ cannot be a proper
subset of any set~$Z$ of atoms that is an elementary loop of
$\Pi_{X,Z}$.
\qed


\subsection{Recognizing Elementary Loops of Disjunctive
  Programs}\label{sec:des-dc}

Although deciding whether a given set of atoms is an elementary loop
of a nondisjunctive program can be done efficiently, it turns out that
the corresponding problem in the case of arbitrary disjunctive programs
is intractable.

\begin{thm}\label{thm:elm-coNP}
For any disjunctive program $\Pi$ and any set~$Y$ of atoms, deciding
whether~$Y$ is an elementary loop of~$\Pi$ is {\sf coNP}-complete.
\end{thm}

\proof
Containment in {\sf coNP} is clear, since it is easy to check that a
given nonempty proper subset of~$Y$ is not outbound in~$Y$ for~$\Pi$.

For {\sf coNP}-hardness, we reduce the {\sf coNP}-hard problem of
deciding whether a finite set~$X$ of atoms is ``unfounded-free'' for
a disjunctive program~$\Pi$ \cite{leo97}, i.e., $X$ contains no
nonempty unfounded sets for~$\Pi$ w.r.t.~$X$. Using a new atom~$e$
that does not occur in~$\Pi$ or~$X$, we construct a program $\Pi'$ as
follows:  for every rule~(\ref{abf-dis}) of~$\Pi_X$, include a rule 
$A \ar e,B,F$ in~$\Pi'$, and, for every $a\in X\cup\{e\}$, include a
rule $e \ar a$ in~$\Pi'$. Given the rules of the latter type, it is
clear that any proper subset~$Z$ of $Y=X\cup\{e\}$ that is not
outbound in~$Y$ for~$\Pi'$ cannot contain~$e$. For every
rule~(\ref{abf-dis}) of~$\Pi'$ such that $A\neq\{e\}$, since $e\in B$,
we then have that
\beq\nonumber
B \cap (Y\setminus Z)\ne \emptyset\ .
\eeq{redu1}
Hence, if a nonempty proper subset $Z$ of $Y$ is not outbound in~$Y$
for~$\Pi'$, for every rule~(\ref{abf-dis}) of~$\Pi'$ such that
$A\neq\{e\}$, at least one of the following conditions holds:
\beq
A \cap Z= \emptyset\ ,
\eeq{redu2}
\beq
A \cap (Y\setminus Z)\ne \emptyset\ ,
\eeq{redu3}
or
\beq
B \cap Z\ne \emptyset\ .
\eeq{redu4}
Since $e\notin A$, (\ref{redu3}) implies that 
\beq
A \cap (X\setminus Z)\ne \emptyset\ .
\eeq{redu5}
We have thus shown that (\ref{redu2}), (\ref{redu4}), or (\ref{redu5})
holds for every rule~(\ref{abf-dis}) of~$\Pi'$ such that $A\neq\{e\}$,
and, similarly, for every rule~(\ref{abf-dis}) of~$\Pi_X$.
Furthermore, we have that
\beq
X\not\models B,F
\eeq{redu6}
for every rule~(\ref{abf-dis}) of~$\Pi\setminus\Pi_X$.
Consequently,
(\ref{redu2}), (\ref{redu4}), (\ref{redu5}), or (\ref{redu6})
holds for every rule~(\ref{abf-dis}) of~$\Pi$, which shows that $Z$ is
unfounded by~$\Pi$ w.r.t.~$X$.
Conversely, if a nonempty subset~$Z$ of~$X$ is unfounded by~$\Pi$
w.r.t.~$X$, the fact that (\ref{redu2}), (\ref{redu4}), or
(\ref{redu5})  holds for every rule~(\ref{abf-dis}) of~$\Pi_X$ implies
that every rule~(\ref{abf-dis}) of~$\Pi'$ satisfies (\ref{redu2}),
(\ref{redu3}), or (\ref{redu4}), so that $Z$ is not outbound in~$Y$
for~$\Pi'$.
Consequently, we conclude that $X$ is unfounded-free for~$\Pi$ iff
$Y=X\cup\{e\}$ is an elementary loop of~$\Pi'$.
\qed

However, for the class of disjunctive programs called
``Head-Cycle-Free'' \cite{ben94}, deciding whether a set of atoms is
an elementary loop is tractable.  We say that a disjunctive
program~$\Pi$ is {\sl Head-Cycle-Free} ({\sl HCF}) if $|A \cap Y|\leq
1$ for every rule~(\ref{abf-dis}) of~$\Pi$ and every loop~$Y$
of~$\Pi$. 

The definition of an {\sl elementary subgraph} for a nondisjunctive
program can be extended to disjunctive programs by modifying the
equation for ${\it EC}_\Pi^{i+1}$ as follows:
\begin{eqnarray*}
{\it EC}_\Pi^{i+1}(X) & = &
\{ (a,b) \mid 
\begin{array}[t]{@{}l@{}}
\text{there is a rule~(\ref{abf-dis}) in $\Pi$ 
       such that $A\cap X = \{a\}$,} \\
\text{$b\in B\cap X$, and all atoms in $B\cap X$ belong to the} \\
\text{same strongly connected component in $(X,{\it EC}^i_\Pi(X))$}\}\ .
\end{array}
\end{eqnarray*}
With this extended definition of an elementary subgraph,
Theorem~\ref{thm:ec-tr} remains correct after replacing 
``nondisjunctive program'' in its statement with 
``HCF program.''

In the next section, we introduce ``Head-Elementary-loop-Free''
programs, and show that Theorem~\ref{thm:ec-tr} can be further
generalized to such programs. 


\section{Head-Elementary-Loop-Free Programs}\label{sec:hef}

In general, computing stable models of a disjunctive program is harder
than computing stable models of a nondisjunctive
program~\cite{eitgot95a}. On the other hand, HCF programs are ``easy''
disjunctive programs that can be turned into equivalent nondisjunctive
programs in polynomial time~\cite{ben94}. This property plays an
important role in the computation of stable models of disjunctive
programs, and is used by answer set solvers {\sc claspd}, 
{\sc cmodels}, and {\sc dlv}. 

By referring to elementary loops in place of loops in the definition
of an HCF program, we define a class of programs that is more general
than HCF programs: we say that a disjunctive program~$\Pi$ is {\sl
  Head-Elementary-loop-Free} ({\sl HEF}) if $|A \cap Y|\leq  1$ for
every rule~(\ref{abf-dis}) of~$\Pi$ and every elementary loop~$Y$
of~$\Pi$.  
Since every elementary loop is also a loop, it is clear that every HCF
program is an HEF program as well. However, not all HEF programs are
HCF.
For example, consider the following program~$\Pi_2$:
\[
\ba l
p \ar r \\
q \ar r \\
r \ar p,q \\
p\ ;\ q \ar\ .
\ea
\]
This program has six loops:  $\{p\}$, $\{q\}$, $\{r\}$, $\{p,r\}$,
$\{q,r\}$, and $\{p, q, r\}$. Since the head of the last rule
contains two atoms from the loop $\{p,q,r\}$, $\Pi_2$ is not HCF. 
On the other hand, $\Pi_2$ is HEF since $\{p,q,r\}$ is not an
elementary loop of~$\Pi_2$: its subsets $\{p,r\}$ and $\{q,r\}$ are
not outbound in~$\{p,q,r\}$ for~$\Pi_2$.

Let us write a rule~(\ref{abf-dis}) in the following form:
\beq
a_1;\dots;a_k\leftarrow B, F \ .
\eeq{rk}
Gelfond {\sl et al.} \citeyear{gel91a} defined a mapping from a
disjunctive program $\Pi$ to a nondisjunctive program
$\Pi_\mathit{sh}$, the {\sl shifted variant} of $\Pi$,
by replacing each rule~(\ref{rk}) with $k>1$ in $\Pi$ by $k$ new rules: 
\beq
a_i\ar B, F,
   \no\ a_1,\dots,\no\ a_{i-1},\no\ a_{i+1},\dots,\no\ a_k\ .
\eeq{rkk}
They showed that every stable model of $\Pi_\mathit{sh}$ is also a
stable model of $\Pi$. Although the converse does not hold in general,
Ben-Eliyahu and Dechter \citeyear{ben94} showed that the converse
holds if $\Pi$ is HCF. We below extend this result to HEF programs.

The following proposition compares the elementary loops of $\Pi$ with
the elementary loops of $\Pi_{\mathit{sh}}$.

\begin{prop}\label{prop:es-hef1}
For any disjunctive program~$\Pi$, if $X$ is an elementary loop
of~$\Pi$, then $X$ is an elementary loop of $\Pi_{\mathit{sh}}$.
\end{prop}

\proof
Assume that~$X$ is an elementary loop of~$\Pi$. Then every nonempty
proper subset~$Y$ of~$X$ is outbound in~$X$ for~$\Pi$, so that there
is a rule~(\ref{rk}) in~$\Pi$ such that
\beq
\{a_1,\dots,a_k\}\cap Y\ne\emptyset\ ,
\eeq{eshef1}
\beq
B\cap (X\setminus Y)\ne\emptyset\ ,
\eeq{eshef2}
\beq\nonumber
\{a_1,\dots,a_k\}\cap (X\setminus Y)=\emptyset\ ,
\eeq{eshef3}
and
\beq
B\cap Y=\emptyset\ .
\eeq{eshef4}
For some $a_i\in\{a_1,\dots,a_k\}\cap Y$, (\ref{eshef1}) implies that 
some rule (\ref{rkk}) in $\Pi_{\mathit{sh}}$ satisfies 
\beq\nonumber
\{a_i\}\cap Y\ne\emptyset
\eeq{eshef5}
and
\beq\nonumber
\{a_i\}\cap (X\setminus Y)=\emptyset\ .
\eeq{eshef6}
Together with~(\ref{eshef2}) and (\ref{eshef4}), this means that $Y$
is outbound in~$X$ for~$\Pi_{\mathit{sh}}$. Consequently, $X$ is an
elementary loop of~$\Pi_{\mathit{sh}}$.
\qed 

The converse of Proposition~\ref{prop:es-hef1} does not hold even if
$\Pi$ is HEF. For example, consider the following HEF program $\Pi_3$:
\[
\ba l
p\ ;\  q\ar r\\
r\ar p\\
r\ar q \ .
\ea
\]
Set $\{p,q,r\}$ is not an elementary loop of~$\Pi_3$ since, for
instance, $\{p\}$ is not outbound in $\{p,q,r\}$ for~$\Pi_3$. On the
other hand, $\{p,q,r\}$ is an elementary loop
of~$(\Pi_3)_\mathit{sh}$:
\beq
\ba l
p\ar r, \no\ q\\
q\ar r, \no\ p\\
r\ar p\\
r\ar q\ .
\ea
\eeq{ex-shift} 

However, the following proposition shows that there is a certain
subset of~$\Pi_\mathit{sh}$ whose elementary loops are also elementary
loops of $\Pi$. 

\begin{prop}\label{prop:es-hef2}
For any disjunctive program~$\Pi$, any set~$X$ of atoms, and any
subset~$Y$ of~$X$, if $Y$ is an elementary loop of $(\Pish)_X$, then
$Y$ is an elementary loop of $\Pi$. 
\end{prop}

\proof
Assume that $Y$ is an elementary loop of $({\Pi_\mathit{sh}})_X$, and
not an elementary loop of $\Pi$ for the sake of contradiction.
Consider any rule (\ref{rk}) in $\Pi$, and any proper subset $Z$ of
$Y$. Since $Y$ is not an elementary loop of $\Pi$, at least one of the
following conditions holds:
\beq
\ba l
\{a_1,\dots,a_k\}\cap Z =\emptyset\ ,  \\
\{a_1,\dots,a_k\}\cap (Y\setminus Z)\neq \emptyset\ , \\
B\cap Z \neq \emptyset\ , \text{ or }\\
B\cap (Y\setminus Z) =\emptyset\ .
\ea
\eeq{eq:es-hef2-1}

We will show that any rule (\ref{rkk}) in $(\Pi_{\mathit{sh}})_X$
obtained from (\ref{rk}) by shifting satisfies at least one of the
following conditions: 
$$
\ba l
\{a_i\}\cap Z =\emptyset\ ,  \\
\{a_i\}\cap (Y\setminus Z)\neq \emptyset\ ,  \\
B\cap Z \neq \emptyset\ , \text{ or } \\
B\cap (Y\setminus Z) =\emptyset\ .
\ea
$$
This contradicts the assumption that $Y$ is an elementary
loop of $(\Pi_{\mathit{sh}})_X$. 

\medskip\noindent
{\sl Case 1:} The first, the third, or the fourth condition of
(\ref{eq:es-hef2-1}) holds. The claim trivially follows.

\medskip\noindent
{\sl Case 2:} $\{a_1,\dots,a_k\}\cap (Y\setminus Z)\ne\emptyset$. 
Recall that 
$$X\models B,\ F,\ not\ a_1,\dots,\ not\ a_{i-1},\ not\ a_{i+1},\
not\ a_k,$$ 
by $({\Pi_\mathit{sh}})_X$ construction.
It follows that $|\{a_1,\dots,a_k\}\cap X|\leq 1$. From the fact that
$Y\subseteq X$ and $Z\subset Y$ we conclude that 
$|\{a_1,\dots,a_k\}\cap Y|\leq 1$ and $\{a_1,\dots,a_k\}\cap Z=\emptyset$, 
so that  $\{a_i\}\cap Z=\emptyset$.
\qed

For instance, for $X=\{p,q,r\}$ and $(\Pi_3)_\mathit{sh}$,
we have that $[(\Pi_3)_\mathit{sh}]_X$ consists of the last two
rules in~(\ref{ex-shift}).  
Only the singletons $\{p\}$, $\{q\}$,
and $\{r\}$ are elementary loops of~$[(\Pi_3)_\mathit{sh}]_X$, and
clearly they are elementary loops of $\Pi_3$ as well.

We are now ready to show the equivalence between an HEF program and
its shifted variant.

\begin{thm}\label{thm:hef-nondisj}
For any HEF program $\Pi$ and any set $X$ of atoms, $X$ is a stable
model of~$\Pi$ iff  $X$ is a stable model of~$\Pi_\mathit{sh}$.
\end{thm}

\proof
{\sl From left to right:} 
Assume that $X$ is a stable model of~$\Pi$. Then $X$ is a model
of~$\Pi_\mathit{sh}$ such that all its atoms occur
in~$\Pi_\mathit{sh}$ and also in~$[\Pish]_X$.
Furthermore, by Theorem$^{\rm d}$~\ref{thm:lf}~(d), we have that $X$
satisfies ${\it LF}_\Pi(Y)$ for all elementary loops~$Y$ of~$\Pi$.
By Proposition~\ref{prop:es-hef2}, the elementary loops of~$\Pi$
include all elementary loops~$Y$ of~$[\Pish]_X$ that are contained
in~$X$, and, since $\Pi$ is HEF, it holds that
${\it ES}_{\Pi_\mathit{sh}}(Y)$  and ${\it ES}_\Pi(Y)$
as well as ${\it LF}_{\Pi_\mathit{sh}}(Y)$ and ${\it LF}_\Pi(Y)$
are equivalent to each other.
This implies that $X$ satisfies ${\it ES}_{\Pi_\mathit{sh}}(Y)$
for all elementary loops~$Y$ of~$[\Pish]_X$ that are contained in~$X$,
so that $X$ contains no elementarily unfounded sets
for~$\Pi_\mathit{sh}$ w.r.t.~$X$. By Theorem$^{\rm
  d}$~\ref{thm:lf}~(e$'$), we conclude that $X$ is a stable model
of~$\Pi_\mathit{sh}$.

\medskip\noindent
{\sl From right to left:}
Assume that $X$ is a stable model of~$\Pi_\mathit{sh}$.
Then $X$ is a model of~$\Pi$ such that all its atoms occur in~$\Pi$.
Furthermore, by Theorem$^{\rm d}$~\ref{thm:lf}~(d), we have that $X$
satisfies ${\it LF}_{\Pi_\mathit{sh}}(Y)$ for all elementary loops~$Y$
of~$\Pi_\mathit{sh}$.
By Proposition~\ref{prop:es-hef1}, the elementary loops
of~$\Pi_\mathit{sh}$ include all elementary loops~$Y$ of~$\Pi$,
and, since $\Pi$ is HEF, it holds that ${\it ES}_\Pi(Y)$ and
${\it ES}_{\Pi_\mathit{sh}}(Y)$ as well as ${\it LF}_\Pi(Y)$ and
${\it LF}_{\Pi_\mathit{sh}}(Y)$ are equivalent to each other.
This implies that $X$ satisfies ${\it LF}_\Pi(Y)$ for all elementary
loops~$Y$ of~$\Pi$. By Theorem$^{\rm d}$~\ref{thm:lf}~(d), we conclude
that $X$ is a stable model of~$\Pi$.
\qed

For instance, one can check that both $\Pi_2$ and
$(\Pi_2)_{\mathit{sh}}$ have $\{p\}$ and $\{q\}$ as their stable
models.
It follows that HEF programs are not more expressive than
nondisjunctive programs, so that one can regard the use of disjunctive
rules in such programs as a syntactic variant.
Furthermore, the problem of deciding whether a model is stable for an
HEF program is tractable, just as the same problem for a
nondisjunctive program. 
(In the case of arbitrary disjunctive programs, it is {\sf
  coNP}-complete \cite{eitgot95a}.)
These properties were known for HCF programs, and here we extended
them to HEF programs.

In Section~\ref{sec:des-dc}, we defined the notion of an elementary
subgraph of a set $X$ of atoms for a disjunctive program~$\Pi$.
Theorem~\ref{thm:ec-tr} still applies to HEF programs.

\setcounter{mythmc}{3}
\addtocounter{mythmc}{-1}
\begin{mythme}
For any HEF program $\Pi$ and any nonempty set~$X$ of atoms
that occur in $\Pi$, $X$ is an elementary loop of~$\Pi$ iff 
the elementary subgraph of $X$ for $\Pi$ is strongly connected.
\end{mythme}

\proof
{\sl From left to right: } 
Assume that $X$ is an elementary loop of~$\Pi$, and, for the sake of
contradiction, that the elementary subgraph of $X$ for~$\Pi$ is not
strongly connected. Then there is a strongly connected component
in~$(X, {\it EC}_\Pi(X))$ whose atoms $Y$ are not reached from any atom in
$X\setminus Y$. Clearly $Y$ is a nonempty proper subset of
$X$, and so is $X\setminus Y$. Furthermore, for every
rule~(\ref{abf-dis}) in~$\Pi$, at least one of the following
conditions holds:
\beq
|A\cap X| > 1\ ,
\eeq{tr0}
\beq
A\cap (X\setminus Y) =\emptyset\ ,
\eeq{tr1}
\beq
B\cap Y= \emptyset\ ,
\eeq{tr2}
or
\beq
B\cap (X\setminus Y)\ne\emptyset\ .
\eeq{tr3}
However, (\ref{tr0}) contradicts the assumption that $\Pi$ is
HEF. Also the fact that at least one of the conditions (\ref{tr1}),
(\ref{tr2}), and (\ref{tr3}) holds contradicts the assumption that 
$X\setminus Y$ is outbound in $X$ for~$\Pi$.

\medskip\noindent
{\sl From right to left: }
Assume that the elementary subgraph of $X$ for~$\Pi$ is strongly
connected. For every nonempty proper subset~$Y$ of~$X$, there is a
minimum integer~$i\geq 0$ such that ${\it EC}_\Pi^i(X)$ does not contain
any edge from an atom in~$Y$ to an atom in~$X\setminus Y$,
but ${\it EC}_\Pi^{i+1}(X)$ contains such an edge.
Thus some rule~(\ref{abf-dis}) in~$\Pi$ satisfies
\beq
|A\cap X| = 1\ ,
\eeq{tr7}
\beq
A\cap Y\ne\emptyset\ ,
\eeq{tr4}
\beq\nonumber
B\cap (X\setminus Y)\neq \emptyset\ ,
\eeq{tr5}
and
\beq\nonumber
B\cap Y=\emptyset\ .
\eeq{tr6}
 From (\ref{tr7}) and (\ref{tr4}), since $Y\subseteq X$,
\beq\nonumber
A\cap (X\setminus Y)=\emptyset\ .
\eeq{tr8}
This shows that $Y$ is outbound in~$X$ for~$\Pi$. We conclude that $X$
is an elementary loop of~$\Pi$.
\qed

Although many properties of HCF programs still apply to HEF programs
(e.g., equivalence between an HEF program and its shifted variant),
the computational complexities of recognizing them are different.
While an HCF program can be recognized in polynomial time
(by computing the strongly connected components of its dependency graph),
Fassetti and Palopoli \citeyear{fas10} showed that deciding whether a
disjunctive program is HEF is {\sf coNP}-complete.\footnote{The
problem was left open in~\cite{gll07}, one of our conference papers
that this paper extends.}
Theorem~\ref{thm:elm-coNP} established a similar complexity gap by
showing that elementary loops are hard to verify in the case of
arbitrary disjunctive programs, while for loops it remains a question of
reachability.

Such elevated complexities may appear daunting, but the semantic
similarities between HEF and HCF programs still exhibit that the
syntactic concept of reachability merely gives a rough approximation
of properties rendering disjunctive programs more difficult than
nondisjunctive ones.
As noted in \cite{fas10}, identifying subclasses of (not necessarily
HCF) disjunctive programs for which verifying the HEF property is
tractable may be an interesting line of future research.

\section{HEF Programs and Inherent Tightness}\label{sec:tight}

When we add more rules to a program, a stable model of the original
program remains to be a stable model of the extended program 
if it satisfies the new rules.

\begin{prop}\label{prop:subset}
For any disjunctive program $\Pi$ and any model $X$ of $\Pi$, $X$ is 
a stable model of~$\Pi$ iff there is a subset $\Pi'$ of $\Pi$ such that
$X$ is a stable model of~$\Pi'$.
\end{prop}

\proof
From left to right is clear.

\medskip\noindent
{\sl From right to left:}
Assume that~$X$ is not a stable model of~$\Pi$. Then some proper
subset~$Y$ of~$X$ is a model of $\Pi^X$. For each subset~$\Pi'$
of~$\Pi$, we have that $(\Pi')^X\subseteq\Pi^X$, so that $Y$ is a
model of $(\Pi')^X$ and $X$ is not a stable model of~$\Pi'$.
\qed 

In view of Theorem$^{\rm d}$~\ref{thm:lf},
Proposition~\ref{prop:subset} tells us that, provided that~$X$ is a
model of~$\Pi$, in order to verify that $X$ is a stable model
of~$\Pi$, it is sufficient to identify a subset~$\Pi'$ of~$\Pi$
such that $X$ is a stable model of~$\Pi'$. Of course, one can
trivially take $\Pi$ itself as the subset $\Pi'$,  but there are
nontrivial subsets that deserve attention. In fact, if $\Pi$ is
nondisjunctive, it is known that the subset $\Pi'$ can be further
restricted to a ``tight'' program \cite{fag94,erd03}---the result
known as ``inherently tight'' \cite{lin03} or ``weakly tight''
\cite{you03} program. In the following, we simplify these notions and
show that they can be extended to HEF programs.

Recall that a loop of $\Pi$ is called {\sl trivial} if it consists of
a single atom such that the dependency graph of $\Pi$ does not contain
an edge from the atom to itself. In other words,  
a loop~$\{a\}$ of~$\Pi$ is trivial if there is no rule~(\ref{abf-dis})
in~$\Pi$ such that $a\in A\cap B$.

\begin{definition} [\cite{lee05}]\label{def:tight}
A disjunctive program $\Pi$ is called {\sl tight} if every loop of
$\Pi$ is trivial.
\end{definition}

As defined previously \cite{apt88,bar94,ino98,lee05}, we call a
set~$X$ of atoms {\sl supported} by a disjunctive program~$\Pi$
if, for every $a\in X$, there is a rule~(\ref{abf-dis}) in $\Pi_X$
such that $A\cap X=\{a\}$.
Note that Definition~\ref{def:tight} and the notion of support also
apply to nondisjunctive programs as a special case.

The property of inherent tightness, introduced by Lin and Zhao
\citeyear{lin03} for the case of nondisjunctive programs,
can now be reformulated and generalized as follows.

\begin{definition}\label{def:inhtight}
A disjunctive program $\Pi$ is called {\sl inherently tight}
on a set~$X$ of atoms if there is a subset $\Pi'$ of $\Pi$ such
that $\Pi'$ is tight and $X$ is supported by $\Pi'$.
\end{definition}

In the case of nondisjunctive programs, this reformulation of inherent
tightness is similar to ``well-supportedness'' \cite{fag94}.
Furthermore, weak tightness, introduced in~\cite{you03}, is closely
related to the notion of inherent tightness.

For nondisjunctive programs, it is known that their stable models can
be characterized in terms of inherent tightness.

\begin{prop} [\cite{fag94,lin03,you03}] \label{prop:inherent-nondis}
For any nondisjunctive program $\Pi$ and any model $X$ of $\Pi$,
$X$ is a stable model of~$\Pi$ iff $\Pi$ is inherently tight on~$X$.
\end{prop}

One may wonder whether Proposition~\ref{prop:inherent-nondis} can be
extended to disjunctive programs as well, given that
Definition~\ref{def:inhtight} readily applies to them.
However, only one direction of Proposition~\ref{prop:inherent-nondis}
holds in the case of arbitrary disjunctive programs.

\begin{prop} \label{prop:inherent-dis}
For any disjunctive program $\Pi$ and any model $X$ of~$\Pi$, if $\Pi$
is inherently tight on~$X$, then $X$ is a stable model of~$\Pi$.
\end{prop}

\proof
Assume that $\Pi$ is inherently tight on $X$. Then there is a
subset~$\Pi'$ of $\Pi$ such that $\Pi'$ is tight and $X$ is supported
by~$\Pi'$. By Proposition~2 from~\cite{lee03a}, $X$ is a stable model
of~$\Pi'$, and, by Proposition~\ref{prop:subset}, $X$ is a stable
model of $\Pi$. 
\qed

To see that the converse of Proposition~\ref{prop:inherent-dis} does
not hold, consider $\Pi$ as follows:
$$
\ba l  
  p \ar q \\
  q \ar p \\
  p \ ;\ q \ar\ .
\ea
$$
Set $\{p,q\}$ is a stable model of $\Pi$. On the other hand,
since any tight subset~$\Pi'$ of~$\Pi$ must exclude the first or the
second rule, it follows that $\{p,q\}$ is not supported by~$\Pi'$.
But this means that $\Pi$ is not inherently tight on $\{p,q\}$.
It is also worthwhile to note that $\{p,q\}$ is an elementary loop
of~$\Pi$, so that $\Pi$ is not HEF (and not HCF).
Indeed, the following theorem tells us that
Proposition~\ref{prop:inherent-nondis} can be extended to HEF
programs.

\begin{thm} \label{thm:inherent-hef}
For any HEF program $\Pi$ and any model $X$ of $\Pi$,
$X$ is a stable model of~$\Pi$ iff $\Pi$ is inherently tight on $X$.
\end{thm}

\proof
{\sl From left to right:}
Assume that $X$ is a stable model of~$\Pi$.
By Proposition$^{\rm d}$~\ref{prop:elm-alt}
(and the fact that every atom of~$X$ occurs in~$\Pi_X$),
any nonempty subset~$Y$ of~$X$ contains some elementary loop~$Z$
of~$\Pi_X$ that is not outbound in~$Y$ for~$\Pi_X$.\footnote{%
If $Y$ is an elementary loop of~$\Pi_X$, take $Z=Y$.}
That is, every rule (\ref{abf-dis}) of~$\Pi_X$ satisfies at least one
of the following conditions:
\beq\nonumber  A\cap Z=\emptyset\ , \eeq{inh1n}
\beq  B\cap (Y\setminus Z)=\emptyset\ , \eeq{inh2}
\beq  A\cap (Y\setminus Z)\ne \emptyset\ , \eeq{inh3}
or
\beq\nonumber  B\cap Z \ne \emptyset\ . \eeq{inh4n} 
 From (\ref{inh3}), since $Y\subseteq X$, 
\beq\nonumber  A\cap (X\setminus Z)\ne \emptyset\ . \eeq{inh6n}
On the other hand, since $X$ is a stable model of~$\Pi$ and $Z$ is a
nonempty subset of $X$, by Theorem$^{\rm d}$~\ref{thm:lf}~(b),
there is a  rule (\ref{abf-dis}) in~$\Pi_X$ such that 
\beq  A\cap Z\ne\emptyset\ , \eeq{inh7}
\beq  A\cap (X\setminus Z)=\emptyset\ , \eeq{inh10}
and
\beq  B\cap Z= \emptyset\ ,  \eeq{inh8} 
so that (\ref{inh2}) must hold,
which, in combination with~(\ref{inh8}), gives us that
\beq  B\cap Y= \emptyset\ .  \eeq{inh11} 
Furthermore, since $\Pi_X\subseteq \Pi$, we have that $Z$ is an
elementary loop of~$\Pi$. Given that $\Pi$ is HEF, from (\ref{inh7})
and (\ref{inh10}), we conclude that 
\beq  A\cap X=\{a\} \eeq{inh12}
for some $a\in Z$, where $a\in Y$ also holds because $Z\subseteq Y$.
We have thus shown that, for any nonempty subset~$Y$ of~$X$,
there is a rule (\ref{abf-dis}) in~$\Pi_X$ such that
(\ref{inh11}) and (\ref{inh12}) for $a\in Y$ hold.
Starting from $X^0=\emptyset$ and $\Pi^0=\emptyset$,
when, for $1\leq i\leq |X|$, we let
\begin{itemize}
\item $Y=(X\setminus X^{i-1})$,
\item $\Pi^i$ is obtained from $\Pi^{i-1}$ by adding some rule
  (\ref{abf-dis}) in~$\Pi_X$ such that (\ref{inh11}) and (\ref{inh12})
  for $a\in Y$ hold, and
\item $X^i=X^{i-1}\cup\{a\}$, 
\end{itemize}
then $X$ is supported by $\Pi'=\Pi^{|X|}$.
Furthermore, since a rule~(\ref{abf-dis}) in $(\Pi^i\setminus\Pi^{i-1})$
satisfies
$A\cap X^{i-1}=\emptyset$ and $B\subseteq X^{i-1}$ 
for every $1\leq i\leq |X|$, $\Pi'$ is tight by construction,
which shows that $\Pi$ is inherently tight on~$X$.

\medskip\noindent
From right to left follows immediately from
Proposition~\ref{prop:inherent-dis}.
\qed

Since every HCF program is HEF, Theorem~\ref{thm:inherent-hef} applies
also to HCF programs.

We demonstrated that, by turning to the notion of an elementary loop
in place of a loop, we obtain generalizations of results known for
loops, such as Theorem~\ref{thm:hef-nondisj}. This brings our
attention to the following question. 
As a tight program can be characterized in terms of loops, can the
notion of a tight program be generalized by referring to elementary
loops instead? To answer it, let us first modify
Definition~\ref{def:tight}  in the following way.

\begin{definition}\label{def:e-tight}
A disjunctive program $\Pi$ is called {\sl e-tight} if every
elementary loop of $\Pi$ is trivial. 
\end{definition}

Since every elementary loop is a loop, it is clear that a tight
program is e-tight as well. But is the class of e-tight programs more
general than the class of tight programs? One reason why this is an
interesting question to consider is because, if so, it would lead to a
generalization of Fages' theorem \cite{fag94}, which would yield a
more general class of programs for which the stable model semantics
coincides with the completion semantics. However, it turns out that
e-tight programs are not more general than tight programs.

\begin{prop}\label{prop:etight}
For any disjunctive program~$\Pi$, $\Pi$ is e-tight iff $\Pi$ is tight.
\end{prop}

\proof
{\sl From left to right:}
Assume that $\Pi$ is not tight.
Then there is a minimal nontrivial loop~$X$ of~$\Pi$,
and the subgraph of the dependency graph of~$\Pi$
induced by~$X$ yields a simple directed cycle. 
That is, for any nonempty proper subset~$Y$ of~$X$,
there is a rule~(\ref{abf-dis}) in~$\Pi$ such that
$A\cap X=\{a\}$, $B\cap X=\{b\}$ for atoms
$a\in Y$, $b\in X\setminus Y$.
This shows that $Y$ is outbound in~$X$ for~$\Pi$,
so that~$X$ is a nontrivial elementary loop of~$\Pi$.

\medskip\noindent
From right to left is clear.
\qed

This result also tells us that the notion of an inherently tight
program does not become more general by referring to elementary loops,
i.e., by replacing the part ``$\Pi'$ is tight'' in the statement of
Definition~\ref{def:inhtight} with ``$\Pi'$ is e-tight.''

\section{HEF Programs and Stability Checking}\label{sec:msv}

For a disjunctive program, the problem of deciding whether a given
model is stable is {\sf coNP}-complete~\cite{eitgot95a}.
On the other hand, in view of Theorem~\ref{thm:hef-nondisj},
the same problem is tractable for HEF programs.
In order to check the stability of a model in polynomial time,
Leone {\sl et al.} \citeyear{leo97} presented an operational
framework, which, for HCF programs, allows for deciding whether a
model is stable.
Given a disjunctive program~$\Pi$ and sets $X$, $Y$ of atoms, they
defined a sequence $R^0_{\Pi,X}(Y),R^1_{\Pi,X}(Y),\dots$, which
converges to a limit $R^\omega_{\Pi,X}(Y)$, in the following way:
\begin{itemize}
\item  $R^0_{\Pi,X}(Y)=Y$ and
\item  $R^{i+1}_{\Pi,X}(Y)$ is obtained by
  removing every atom $a$ from $R^i_{\Pi,X}(Y)$ 
  such that some rule (\ref{abf-dis}) in $\Pi_X$
  satisfies
  $A\cap (X\cup\{a\})=\{a\}$ and $B\cap R^i_{\Pi,X}(Y)=\emptyset$.\footnote{%
If $Y$ is a subset of~$X$,
the condition ``$A\cap (X\cup\{a\})=\{a\}$'' can be replaced with
``$A\cap X=\{a\}$'' without altering $R^\omega_{\Pi,X}(Y)$.}
\end{itemize}

The disjunctive rules considered in \cite{leo97} do not admit double
negations in rule bodies, but its (dis)use merely affects conditions
like $X\models B,F$ (or $X\not\models B,F$) determining $\Pi_X$, while
it is inconsequential otherwise. Hence, the following results remain
valid for disjunctive programs whose rules are of the
form~(\ref{abf-dis}).

\begin{prop}[{\cite[Lemma~6.4]{leo97}}]\label{prop:hef-unfsub}
For any disjunctive program~$\Pi$ and any sets $X$, $Y$ of atoms
that occur in~$\Pi$, all subsets of~$Y$ that are unfounded by~$\Pi$
w.r.t.~$X$ are contained in $R^\omega_{\Pi,X}(Y)$.
\end{prop}

\begin{prop}[{\cite[Proposition~6.5]{leo97}}]\label{prop:hef-unfounded}
For any disjunctive program~$\Pi$ and any set $X$ of atoms
that occur in~$\Pi$, if $R^\omega_{\Pi,X}(X)=\nolinebreak\emptyset$,
then $X$ contains no nonempty unfounded sets for~$\Pi$ w.r.t.~$X$.
\end{prop}

\begin{prop}[{\cite[Theorem~6.9]{leo97}}]\label{prop:hef-check}
For any HCF program~$\Pi$ and any set $X$ of atoms
that occur in~$\Pi$,
$X$ contains no nonempty unfounded sets for~$\Pi$ w.r.t.~$X$
iff $R^\omega_{\Pi,X}(X)=\emptyset$.
\end{prop}

For a model~$X$ of~$\Pi$, 
in view of Theorem$^{\rm d}$~\ref{thm:lf}~(b$'$),
Proposition~\ref{prop:hef-unfounded} tells us that $X$ is a stable
model of~$\Pi$ if $R^\omega_{\Pi,X}(X)=\emptyset$.
As stated in Proposition~\ref{prop:hef-check}, the converse also holds
if $\Pi$ is HCF. We below extend this result to HEF programs.

\begin{prop}\label{prop:hef-relem}
For any HEF program~$\Pi$, any set $X$ of atoms, and any subset~$Y$
of~$X$ whose atoms occur in~$\Pi$, if $R^\omega_{\Pi,X}(Y)\ne\emptyset$,
then $R^\omega_{\Pi,X}(Y)$ contains an elementary loop~$Z$
of~$\Pi$ that is unfounded by~$\Pi$ w.r.t.~$X$.
\end{prop}

\proof
Assume that $R^\omega_{\Pi,X}(Y)\ne\emptyset$.
Then, for every rule~(\ref{abf-dis}) of~$\Pi_X$, at least one of the
following conditions holds: 
\beq
|A\cap X|>1\ ,
\eeq{rel1}
\beq
A\cap R^\omega_{\Pi,X}(Y)=\emptyset\ ,
\eeq{rel2}
or
\beq
B\cap R^\omega_{\Pi,X}(Y)\ne\emptyset\ .
\eeq{rel3}
By Proposition$^{\rm d}$~\ref{prop:elm-alt}, $R^\omega_{\Pi,X}(Y)$
contains some elementary loop~$Z$ of~$\Pi$
that is not outbound in~$R^\omega_{\Pi,X}(Y)$ for~$\Pi$.\footnote{%
If $R^\omega_{\Pi,X}(Y)$ is an elementary loop of~$\Pi$, take $Z=R^\omega_{\Pi,X}(Y)$.}
For the sake of contradiction, assume that $Z$ is not unfounded by~$\Pi$ w.r.t.~$X$.
Then there is a rule (\ref{abf-dis}) in~$\Pi_X$ such that
\beq  A\cap Z\ne\emptyset\ , \eeq{rel4}
\beq  A\cap (X\setminus Z)= \emptyset\ , \eeq{rel5}
and
\beq  B\cap Z = \emptyset\ . \eeq{rel6} 
 From (\ref{rel5}), since $R^\omega_{\Pi,X}(Y)\subseteq X$, 
\beq  A\cap (R^\omega_{\Pi,X}(Y)\setminus Z)= \emptyset\ , \nonumber\eeq{rel7}
which, in combination with~(\ref{rel4}), (\ref{rel6}),
and the fact that~$Z$ is not outbound in $R^\omega_{\Pi,X}(Y)$ for~$\Pi$,
gives us that
\beq  B\cap (R^\omega_{\Pi,X}(Y)\setminus Z)=\emptyset\ . \eeq{rel8}
From~(\ref{rel6}) and~(\ref{rel8}),
we conclude that (\ref{rel3}) does not hold.
Furthermore, since \hbox{$Z\subseteq R^\omega_{\Pi,X}(Y)$},
(\ref{rel4}) implies that (\ref{rel2}) does not hold.
Hence, (\ref{rel1}) must hold,
which, in combination with~(\ref{rel5}), gives us that
\beq
|A\cap Z|>1\ .
\nonumber\eeq{rel9}
But since $Z$ is an elementary loop of~$\Pi$, this contradicts that
$\Pi$ is HEF.
\qed

We are now ready to generalize Proposition~\ref{prop:hef-check}
to HEF programs.

\begin{thm}\label{thm:hef-check}
For any HEF program~$\Pi$ and any set $X$ of atoms that occur in~$\Pi$,
$X$ contains no nonempty unfounded sets for~$\Pi$ w.r.t.~$X$ iff
$R^\omega_{\Pi,X}(X)=\emptyset$.
\end{thm}

\proof
From left to right follows immediately from
Proposition~\ref{prop:hef-relem}.

\medskip\noindent
From right to left follows immediately from
Proposition~\ref{prop:hef-unfounded}.
\qed

Regarding the models of HEF programs, we derive the following
corollary. 

\begin{cor}\label{col:hef-check}
For any HEF program~$\Pi$ and any set $X$ of atoms
that occur in~$\Pi$,
$X$ is a stable model of~$\Pi$
iff 
$X$ is a model of~$\Pi$ such that $R^\omega_{\Pi,X}(X)=\emptyset$.
\end{cor}
\proof
Both directions follow immediately from 
Theorem$^{\rm d}$~\ref{thm:lf}~(b$'$) and
Theorem~\ref{thm:hef-check}.
\qed

For instance, reconsider $\Pi_2$ from Section~\ref{sec:hef}, which is
HEF, but not HCF. 
Hence,  Corollary~\ref{col:hef-check} applies, but
Proposition~\ref{prop:hef-check} does not apply.
Indeed, since $R^\omega_{\Pi_2,X}(X)=X$
for (non-stable) model $X=\{p,q,r\}$ of~$\Pi_2$,
Corollary~\ref{col:hef-check} allows us to conclude that $X$
is not a stable model of~$\Pi_2$.
On the other hand, for model $\{p\}$ of~$\Pi_2$,
we have that $R^\omega_{\Pi_2,\{p\}}(\{p\})=\nolinebreak\emptyset$,
which implies that $\{p\}$ is a stable model of~$\Pi_2$.

In Section~\ref{sec:eusdp}, we defined the notion of an elementarily
unfounded set and showed that it coincides with a minimal nonempty
unfounded set. Thus stability checking can be cast into the problem
of ensuring the absence of elementarily unfounded sets. Since every
elementarily unfounded set is a loop, it is clearly contained in a
maximal loop, which allows us to modularize the consideration of
(elementarily) unfounded sets. The idea of using maximal loops for
partitioning a program and confining stability checking to subprograms
was already exploited by Leone {\sl et al.} \citeyear{leo97} and Koch
{\sl et al.} \citeyear{koc03}. In fact, for a disjunctive
program~$\Pi$ and a set~$X$ of atoms, Leone {\sl et al.}
\citeyear{leo97} showed how stability can be checked separately for
maximal loops of~$\Pi$, and Koch {\sl et al.} \citeyear{koc03}
developed this idea further by considering maximal loops of the
smaller program $\Pi_{X,R^\omega_{\Pi,X}(X)}$.
We below describe a notion  called ``bounding loops,''
which go beyond such maximal loops. 

For a disjunctive program~$\Pi$ and a set~$X$ of atoms, 
we say that a subset~$Y$ of~$X$ is a {\sl bounding loop} of~$\Pi$
w.r.t.~$X$ if $Y$ is maximal among all subsets~$Z$ of~$X$
such that $Z$ is a loop of $\Pi_{X,Z}$ and $R^\omega_{\Pi,X}(Z)=Z$.
Note that there are two crucial differences between bounding loops and
elementarily unfounded sets.
First, a bounding loop~$Y$ of~$\Pi$ w.r.t.~$X$ is not necessarily an
elementary loop of $\Pi_{X,Y}$.
Next, it does not need to be unfounded by~$\Pi$ w.r.t.~$X$.
To see this,
observe that $X=\{p,q,r\}$ is a bounding loop of~$\Pi_2$ w.r.t.~$X$
that is not (elementarily) unfounded by~$\Pi_2$ w.r.t.~$X$.
Furthermore, one can check that $\{p,r\}$ and $\{q,r\}$
are (elementarily) unfounded by~$\Pi_2$ w.r.t.~$X$,
and thus Proposition$^{\rm d}$~\ref{prop:max-elm}
tells us that $X$ is not an elementary loop of $(\Pi_2)_{X,X}$.

The following two propositions describe properties of bounding loops
that are similar to those of maximal loops, as used in~\cite{leo97,koc03}.

\begin{prop}\label{prop:bounding-disj}
For any disjunctive program~$\Pi$ and any set~$X$ of atoms, all
bounding loops of~$\Pi$ w.r.t.~$X$ are mutually disjoint.
\end{prop}

\proof  
Let $Y_1$, $Y_2$ be subsets of~$X$ such that
$Y_1$ is a loop of $\Pi_{X,Y_1}$,
$Y_2$ is a loop of $\Pi_{X,Y_2}$,
$R^\omega_{\Pi,X}(Y_1)=Y_1$, and
$R^\omega_{\Pi,X}(Y_2)=Y_2$.
If $Y_1\cap Y_2\ne\emptyset$,
the fact that $\Pi_{X,Y_1}\cup\Pi_{X,Y_2}\subseteq \Pi_{X,Y_1\cup Y_2}$
implies that
$Y_1\cup Y_2$ is a loop of $\Pi_{X,Y_1\cup Y_2}$.
Furthermore,
since
$R^\omega_{\Pi,X}(Y_1)=Y_1$ and
$R^\omega_{\Pi,X}(Y_2)=Y_2$,
for any rule~(\ref{abf-dis}) of~$\Pi_X$ such that 
\hbox{$A\cap(Y_1\cup Y_2)\ne\emptyset$} and
$B\cap(Y_1\cup Y_2)=\emptyset$,
we have that $|A\cap X|> 1$, from which we conclude that 
$R^\omega_{\Pi,X}(Y_1\cup Y_2)=Y_1\cup Y_2$.
Since bounding loops of~$\Pi$ w.r.t.~$X$ are maximal among all
subsets~$Z$ of~$X$ such that $Z$ is a loop of $\Pi_{X,Z}$ and
$R^\omega_{\Pi,X}(Z)=Z$, this shows that they must be mutually disjoint.
\qed

\begin{prop}\label{prop:eus-bounding}
For any disjunctive program~$\Pi$ and any set $X$ of atoms
that occur in~$\Pi$, every elementarily unfounded set for~$\Pi$
w.r.t.~$X$ is a singleton or contained in some bounding loop of~$\Pi$
w.r.t.~$X$.
\end{prop}

\proof
Assume that $Y$ is an elementarily unfounded set for~$\Pi$ w.r.t.~$X$
that is not a singleton. From the definition of an elementarily
unfounded set, it follows that $Y$ is a loop of $\Pi_{X,Y}$ and a
subset of~$X$. Since $Y$ is also unfounded by~$\Pi$ w.r.t.~$X$, by
Proposition~\ref{prop:hef-unfsub}, we have that $R^\omega_{\Pi,X}(Y)=Y$.
This shows that $Y$ is contained in some maximal subset~$Z$ of~$X$
such that $Z$ is a loop of $\Pi_{X,Z}$ and $R^\omega_{\Pi,X}(Z)=Z$.
\qed

Proposition~\ref{prop:bounding-disj} and Proposition~\ref{prop:eus-bounding}
tell us that checking the absence of elementarily
unfounded sets can be accomplished separately for bounding loops.

\begin{prop}\label{prop:stable-nobounding}
For any disjunctive program $\Pi$ and any model $X$ of~$\Pi$,
$X$ is a stable model of~$\Pi$ iff
\begin{itemize}
\item  $X$ is supported by~$\Pi$  and
\item  no bounding loop of~$\Pi$ w.r.t.~$X$ 
       contains a nonempty unfounded set for~$\Pi$ w.r.t~$X$.
\end{itemize}
\end{prop}

\proof
 From left to right follows immediately from 
Theorem$^{\rm d}$~\ref{thm:lf}~(b$'$)
(and the fact that every atom of~$X$ occurs in~$\Pi$). 

\medskip\noindent
{\sl From right to left:}
Assume that $X$ is not a stable model of~$\Pi$ but supported by~$\Pi$.
Then, by Theorem$^{\rm d}$~\ref{thm:lf}~(e$'$),
$X$ contains some elementarily unfounded set~$Y$ for~$\Pi$ w.r.t.~$X$.
If $Y$ is not a singleton,
by Proposition~\ref{prop:eus-bounding},
$Y$ is contained in some bounding loop of~$\Pi$ w.r.t.~$X$.
Otherwise, if $Y$ is a singleton,
the assumption that $X$ is supported by~$\Pi$
implies that there is a rule~(\ref{abf-dis}) in $\Pi_{X,Y}$
such that $A\cap X=Y$,
so that $Y$ is a loop of~$\Pi_{X,Y}$.
Since $Y$ is unfounded by~$\Pi$ w.r.t.~$X$,
by Proposition~\ref{prop:hef-unfsub}, we also have that
$R^\omega_{\Pi,X}(Y)=Y$.
This shows that $Y$ is contained in some
maximal subset~$Z$ of~$X$ such that $Z$ is a loop of $\Pi_{X,Z}$ and
$R^\omega_{\Pi,X}(Z)=Z$.
\qed

As it is easy to check that~$X$ is supported by~$\Pi$,
Proposition~\ref{prop:stable-nobounding} tells us that the
investigation of bounding loops constitutes the hard part of stability
checking.
But this is not due to the hardness of identifying them. 
In fact, the following method can be used to compute
all bounding loops~$Z$ of~$\Pi$ w.r.t.~$X$ in polynomial time:
\begin{enumerate}
\item Let $Y=X$.
\item Let $Z=R^\omega_{\Pi,X}(Y)$. 
      \hfill (Note that $Z=R^\omega_{\Pi,X}(Z)$.)
 \item If $Z$ is a loop of~$\Pi_{X,Z}$, then mark $Z$ as a bounding loop
        of~$\Pi$ w.r.t.~$X$.
       Otherwise, proceed with Step 2
        for every maximal loop~$Y$ of $\Pi_{X,Z}$ that is contained in~$Z$.
\end{enumerate}
The soundness of this approach is straightforward,
given that operator~$R$ is monotone,
i.e., $R^\omega_{\Pi,X}(Z) \subseteq R^\omega_{\Pi,X}(Y)$ if $Z \subseteq Y$
(used in Step~2),
and likewise that $\Pi_{X,Y}\subseteq \Pi_{X,Z}$ if $Y \subseteq Z$
(used in Step~3).

For illustration, consider the following program~$\Pi_4$:
\[
\ba {r@{}c@{}l@{\hspace{16mm}}r@{}c@{}l@{\hspace{16mm}}r@{}c@{}l@{\hspace{14mm}}l}
p & {}\ar{} & r       &
s & {}\ar{} & p &
p \ ;\ q & {}\ar{} & u \\
q & {}\ar{} & r       &
s & {}\ar{} & t       &
s \ ;\ t & {}\ar{} & q 
\\
r & {}\ar{} & p,q     &
t & {}\ar{} & s,u      &
r \ ;\ u & {}\ar{} & t &
u \ ;\ v \ar \ .
\ea
\]
For $X=\{p,q,r,s,t,u\}$, it holds that $(\Pi_4)_{X,X}=\Pi_4$ and 
also that $X$ is a loop of~$\Pi_4$.
However, $Y=R^\omega_{\Pi_4,X}(X)=\{p,q,r,s,t\}\neq X$, so that $X$
is not a bounding loop of~$\Pi_4$ w.r.t.~$X$.
On the other hand, $Y$ is not a loop of $(\Pi_4)_{X,Y}$,
which does not include the last two rules where $u$ occurs in the
head. Rather, $Z=\{p,q,r\}$ and $Z'=\{s,t\}$ are the maximal loops of
$(\Pi_4)_{X,Y}$ that are contained in~$Y$.
In view of the rules in the second column,
$R^\omega_{\Pi_4,X}(Z')=\emptyset$,
which shows that no subset of~$Z'$ is a bounding loop of~$\Pi_4$
w.r.t.~$X$.
For~$Z=\{p,q,r\}$, we obtain $R^\omega_{\Pi_4,X}(Z)=Z$, and
$Z$ is also a loop of $(\Pi_4)_{X,Z}$
(due to the rules in the first column).
That is, $Z$ is the single bounding loop of~$\Pi_4$ w.r.t.~$X$.

Let us compare this outcome with the ones 
of previous methods for modularizing stability checking.
Since the approach described in \cite{leo97}
considers maximal loops of the original program and
$X=\{p,q,r,s,t,u\}$ is a (maximal) loop of~$\Pi_4$,
it cannot be used to decompose~$X$, and the only
applicable simplification is to remove~$u$ 
by means of $R$, that is, $Y=R^\omega_{\Pi_4,X}(X)=\{p,q,r,s,t\}$.
The approach in \cite{koc03}
considers the maximal loops of $\Pi_{X,R^\omega_{\Pi,X}(X)}$,
which, for $(\Pi_4)_{X,Y}$,
gives $Z=\{p,q,r\}$ and $Z'=\{s,t\}$.
As described in \cite{koc03}, since the subprograms
$(\Pi_4)_{X,Z}$ and $(\Pi_4)_{X,Z'}$ are not HCF,
they are not simplified any further and used to separately check for
a nonempty unfounded subset of~$Z$ or~$Z'$, respectively.
Unlike this, the notion of a bounding loop allowed us to 
eliminate all subsets of~$Z'$ as potential nonempty unfounded sets.

Turning back to HEF programs, from Proposition~\ref{prop:hef-relem},
we derive the following proposition for a subprogram $\Pi_{X,Y}$
associated with a bounding loop~$Y$ of~$\Pi$ w.r.t.~$X$.

\begin{prop}\label{prop:hef-subunfounded}
For any disjunctive program $\Pi$, any set $X$ of atoms, and any
bounding loop~$Y$ of~$\Pi$ w.r.t.~$X$,
if $\Pi_{X,Y}$ is HEF, 
then $Y$ contains a nonempty unfounded set for~$\Pi$
w.r.t.~$X$.
\end{prop}

\proof
Assume that $\Pi_{X,Y}$ is HEF.
 From the definition of a bounding loop,
it follows that all atoms of~$Y$ occur in $\Pi_{X,Y}$ and that
$Y=R^\omega_{\Pi,X}(Y)=R^\omega_{(\Pi_{X,Y}),X}(Y)\subseteq\nolinebreak X$.
By Proposition~\ref{prop:hef-relem},
we conclude that
$R^\omega_{(\Pi_{X,Y}),X}(Y)$ contains an elementary loop~$Z$
of~$\Pi_{X,Y}$ that is unfounded by~$\Pi_{X,Y}$ w.r.t.~$X$.
From the definition of~$\Pi_{X,Y}$
and since $Z$ is contained in~$Y$,
we conclude that~$Z$ is also unfounded by~$\Pi$ w.r.t.~$X$.
\qed

Proposition~\ref{prop:hef-subunfounded} tells us that the existence of
a bounding loop~$Y$ of~$\Pi$ w.r.t.~$X$ whose associated
subprogram~$\Pi_{X,Y}$ is HEF is already sufficient to conclude
that~$X$ is not a stable model of~$\Pi$. Reconsidering 
the bounding loop $Z=\{p,q,r\}$ of~$\Pi_4$
w.r.t.~$X=\{p,q,r,\linebreak[1]s,t,u\}$,
we have that $(\Pi_4)_{X,Z}$, consisting of the rules in the first
column along with the disjunctive rule containing $p$ and~$q$ in the
head, is HEF (neither $\{p,q\}$ nor $\{p,q,r\}$ is an elementary loop
of~$(\Pi_4)_{X,Z}$).
Thus $X$ is not a stable model of~$\Pi_4$.
Indeed, $Z$ contains two (elementarily) unfounded sets for~$\Pi_4$
w.r.t.~$X$: $\{p,r\}$ and $\{q,r\}$.

\section{Conclusion}\label{sec:conclusion}

The notion of an elementary loop and its properties provide useful
insights into the concept of a loop and the relationship between
nondisjunctive programs and disjunctive programs.
By turning to the notion of an elementary loop in place of a loop, 
we could strengthen the theorem by Lin and Zhao \citeyear{lin04},
its generalization to disjunctive programs \cite{lee03a}, and
the main theorem (for programs in canonical form) from \cite{lee05}.
The semantic (e.g., Theorem$^{\rm d}$~\ref{thm:min-uf})
and complexity-theoretic (e.g., Theorem~\ref{thm:elm-coNP})
properties of elementary loops indicate their close relationship
to unfounded sets. Compared with loops, elementary loops provide a
deeper understanding of the internal structure of unfounded sets.
In fact, we have shown that loop formulas of non-elementary loops are
unnecessary for checking the stability of a model. 
It is an interesting open question whether this result can be used
to improve computation performed by answer set solvers
that calculate loop formulas,
such as 
{\sc assat},
{\sc claspd}, and
{\sc cmodels}.

A method to identify an elementarily unfounded set for an HEF program
was presented in \cite{gll07}.\footnote{%
It is omitted in this paper for brevity.}
An orthogonal approach is implemented in {\sc claspd}: it greedily
adds atoms~$a$ to an unfounded set~$Y$ if $Y\cup\{a\}$ stays unfounded
and has a shorter loop formula than~$Y$.
However, the potential of unfounded set ``optimization''
has not been studied in-depth so far, and the theoretical foundations
laid here may be useful for future investigations in this direction.
Regarding nondisjunctive programs, elementary loops can be distinct
from loops of programs called ``binary'' \cite{janhunen06a}.
Moreover, modularity aspects of disjunctive stable models
\cite{jaoitowo09a}, which are closely related to loop formulas
(Lemma~5.4 from \cite{jaoitowo09a}),
can be refined by referring to elementary loops in place of loops.
Lifting elementary loops to first-order programs,
as already done for loops \cite{chliwaza06a,lee08a},
may also be a direction to explore.

The notion of an HEF program is a strict generalization of the notion
of an HCF program. On the one hand, nice properties of HCF programs
still apply to HEF programs.
In particular, their inherent tightness on stable models grants the
soundness of shifting head atoms into the body as well as
the possibility of performing stability checks in polynomial time.
This however implies that a disjunctive program encoding an instance of a
$\Sigma^P_2$-hard problem is unlikely to be HEF.
On the other hand, recognizing elementary loops and verifying the HEF
property are both intractable in the case of arbitrary disjunctive
programs, but tractable for nondisjunctive and HCF programs.
This parallels the complexity of stability checking \cite{eitgot95a},
and it also tells us that the inherent complexities of computational tasks
dealing with elementary loops tightly correlate to the fragment of
disjunctive programs under consideration.
As the latter does not apply to recognizing loops or verifying the HCF
property, the notion of an HEF program more precisely renders what makes
arbitrary disjunctive programs more difficult than nondisjunctive programs.
Whether this admits (syntactic) characterizations of yet unknown
subclasses of disjunctive programs for which verifying the HEF
property is tractable is an interesting open question.

\section*{Acknowledgments}
We are grateful to Selim Erdo\u gan, Tomi Janhunen, Dan Lessin,
Vladimir Lifschitz, Torsten Schaub, Jicheng Zhao, and the anonymous
referees of \cite{gll06,gll07} and this paper for their useful
comments. 
Martin Gebser was partially supported by the German Research Foundation
under Grant \mbox{SCHA~550/8-1}.
Joohyung Lee was partially supported by the National Science
Foundation under Grant \mbox{IIS-0916116} and by the Office of the
Director of National Intelligence (ODNI), Intelligence Advanced
Research Projects Activity (IARPA), through US army.
Yuliya Lierler was partially supported by the National Science
Foundation under Grant \mbox{IIS-0712113} and by a 2010 Computing
Innovation Fellowship.

\bibliographystyle{acmtrans}

\end{document}